\documentclass[journal]{IEEEtran}
\usepackage{cite}
\usepackage{support-caption}
\usepackage{subcaption}
\usepackage[utf8]{inputenc}
\usepackage{caption}
\usepackage{multirow}
\usepackage{blindtext}
\usepackage{tabularx}
\usepackage[table,xcdraw]{xcolor}
\usepackage{array}
\usepackage{soul}
\usepackage{color}
\usepackage{booktabs}
\usepackage{times}
\usepackage{soul}
\usepackage{url}
\usepackage[hidelinks]{hyperref}
\usepackage{amsmath}
\usepackage{graphicx}
\usepackage{xspace}
\usepackage{fancyhdr}
\usepackage{graphics}
\usepackage{makecell}
\usepackage{csquotes}
\usepackage{xr}
\usepackage{amssymb}
\usepackage{tikz}
\usepackage{pgfplots}
\pgfplotsset{compat=1.16}
\usepackage{wrapfig}
\usepackage{placeins}
\usepackage{subcaption} 
\usepackage{enumitem}
\usepackage{ulem} 
\usepackage{balance}
\usepackage[T1]{fontenc}
\definecolor{green}{rgb}{0.0, 0.5, 0.0}
\definecolor{amethyst}{rgb}{0.6, 0.4, 0.8}
\usepackage{colortbl}

\begin{document}
\title{DVFL-Net: A Lightweight Distilled Video Focal Modulation Network for Spatio-Temporal Action Recognition}

\author{Hayat Ullah, Muhammad Ali Shafique, Abbas Khan, 
Arslan Munir,~\IEEEmembership{Senior Member,~IEEE,}
\thanks{Hayat Ullah, Abbas Khan, and Arslan Munir are with the Intelligent Systems, Computer Architecture, Analytics, and Security Laboratory (ISCAAS Lab), Department of Electrical Engineering and Computer Science, Florida Atlantic University, Boca Raton, FL 33431, USA (e-mail: hullah2024@fau.edu, abbaskhan2024@fau.edu, arslanm@fau.edu).}
\thanks{Muhammad Ali Shafique is with the Intelligent Systems, Computer Architecture, Analytics, and Security Laboratory (ISCAAS Lab), Department of Electrical and Computer Engineering, Kansas State University, Manhattan, KS 66506, USA (e-mail: alishafique@ksu.edu).}
}
\maketitle

\begin{abstract}
The landscape of video recognition has undergone a significant transformation, shifting from traditional Convolutional Neural Networks (CNNs) to Transformer-based architectures in order to achieve better accuracy. While CNNs, especially 3D variants, have excelled in capturing spatiotemporal dynamics for action recognition, recent developments in Transformer models, with their self-attention mechanisms, have proven highly effective in modeling long-range dependencies across space and time. Despite their state-of-the-art performance on prominent video recognition benchmarks, the computational demands of Transformers, particularly in processing dense video data, remain a significant hurdle. To address these challenges, we introduce a lightweight Video Focal Modulation Network named DVFL-Net, which distills the spatio-temporal knowledge from a large pre-trained teacher to nano student model, making it well-suited for on-device applications. By leveraging knowledge distillation and spatial-temporal feature extraction, our model significantly reduces computational overhead (approximately 7$\times$) while maintaining high performance on video recognition tasks.
We combine the forward Kullback–Leibler (KL) divergence and spatio-temporal focal modulation to distill the local and global spatio-temporal context from the Video-FocalNet Base (teacher) to our proposed nano VFL-Net (student) model. We extensively evaluate our DVFL-Net, both with and without forward KL divergence, against recent state-of-the-art HAR approaches on UCF50, UCF101, HMDB51, SSV2 and Kinetics-400 datasets. Further, we conducted a detailed ablation study in forward KL divergence settings and report the obtained observations.  The obtained results confirm the superiority of the distilled VFL-Net (i.e., DVFL-Net) over existing methods, highlighting its optimal tradeoff between performance and computational efficiency, including reduced memory usage and lower GFLOPs, making it a highly efficient solution for HAR tasks. \href{https://github.com/iscaas/AFOSR-HAR-2021-2025/tree/main/DVFL-Net}{\textcolor{purple}{\texttt{https://github.com/iscaas/DVFL-Net}}.}

\end{abstract}
\IEEEpeerreviewmaketitle
\section{Introduction}
\IEEEPARstart{T}{he} evolution of video recognition technologies has seen a significant shift from traditional Convolutional Neural Networks (CNNs) to Transformer-based architectures. This transition is driven by the need for greater accuracy in video processing. CNNs have long been the backbone of video recognition, leveraging their ability to extract spatial features effectively. However, recent advancements in Transformer models have demonstrated superior performance on major video recognition benchmarks, indicating a promising future for these architectures in video action recognition domain. 

CNNs, particularly 3D CNNs~\cite{carreirajoao2017, christophr2016, trandu2015}, have shown remarkable capabilities in action recognition tasks by capturing spatiotemporal features directly from video frames. tudies indicate that 3D CNNs outperform 2D CNNs~\cite{karpathytoderici2014, yuejoe2015, simonyankaren2014} counterparts by effectively modeling the temporal dynamics of video data, which is crucial for understanding actions~\cite{anvarov2020,tran2018}. Additionally, two-stream CNN architectures, which utilize separate streams for spatial and temporal information, have been widely adopted, achieving competitive accuracy across various datasets~\cite{sarabu2021, kim2020}.
\begin{figure}[t]
	\centering
	\includegraphics[width=\linewidth]{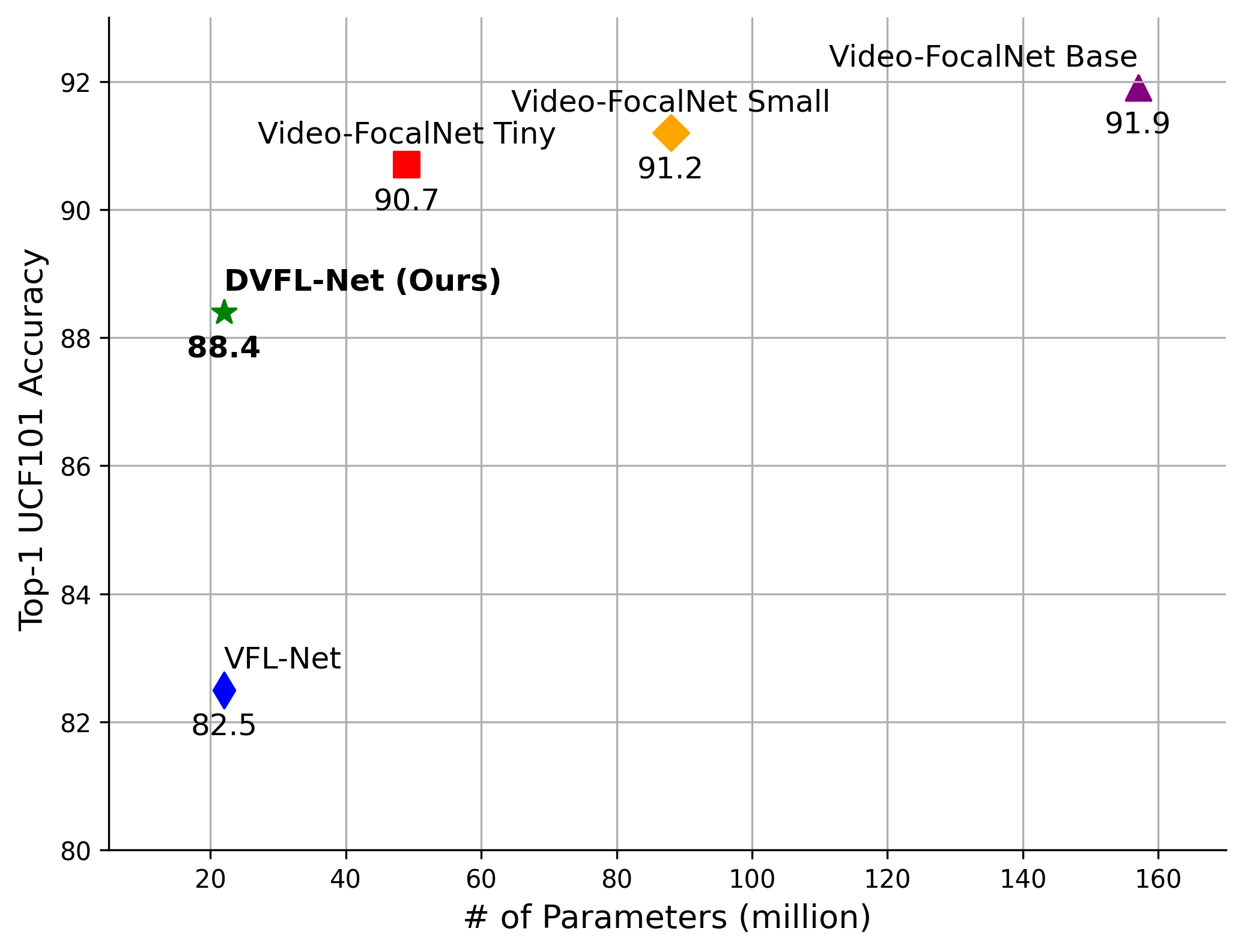}
	\caption{\textbf{\textit{Model's Top-1 Accuracy vs Model's Parameters Cost}}: We evaluate the performance of our DVFL-Net model against different variants of the Video-FocalNet \cite{wasim2023video} architecture for video action recognition task. The comparison is based on top-1 accuracy vs model’s parameters cost, using the \textbf{UCF101} dataset. Our DVFL-Net demonstrates competitive performance compared to its Video-FocalNet counterparts.}
	\label{fig:acc_vs_complexity}
\end{figure}

On the other hand, Transformer models have introduced a new paradigm by leveraging self-attention~\cite{vaswani2017attention} mechanisms that capture long-range dependencies across both spatial and temporal dimensions~\cite{hedgeagrawal2018,liuning2021} and have proven highly effective for video recognition~\cite{gberta2021timesformer, arnab2021vivit, liu2021video-swin, shen2022mtv}. These models have been shown to achieve better state-of-the-art performance than traditional CNNs, for example, the Video Swin Transformer~\cite{hedgeagrawal2018} achieves higher accuracy than CNNs due to its stronger locality inductive bias. Furthermore, Studies suggest that Transformers outperform semi-supervised CNN methods, particularly in data-scarce scenarios~\cite{arnabdehghani2021}.

The comparative analysis of CNNs and Transformers reveals that while CNNs excel in extracting local features in video recognition, Transformers are emerging as an alternative due to their ability to capture global context and relationships across video frames. This allows ViTs to generalize better to large datasets, as shown by recent results on video recognition benchmarks~\cite{kay2017k400, carreira2019k700, goyal2017ssv2} where they have outperformed their CNN counterparts.

However, the computational expense associated with processing dense video frames remains a challenge for ViTs. This is a significant concern, primarily due to the inherent complexity of self-attention mechanisms that scale quadratically with the number of tokens. This challenge arises because video data is typically represented as a sequence of frames, which significantly increases the number of tokens. For instance, Wang et al. highlight that, although video transformers achieve remarkable performance, they incur high computational costs due to the necessity of calculating attention across all tokens in a video sequence~\cite{wang2023tsnet}. Furthermore, Liang et al. discuss how traditional transformer architectures struggle with the high computational demands of self-attention on large sets of 3D tokens, which can hinder their efficiency in video recognition tasks~\cite{liang2021dual}.\\
\indent To tackle these challenges, we thoroughly explore different design configurations of the Video-FocalNet architecture \cite{wasim2023video} (which serves as the foundation of our proposed method and the baseline for this research) and develop a computationally efficient VFL-Net model, optimized for spatio-temporal context modeling using nano-scale spatio-temporal focal modulation mechanism. Our analysis demonstrates that the proposed VFL-Net achieves an optimal tradeoff between model performance and computational efficiency. Additionally, we leverage knowledge distillation to transfer rich spatio-temporal semantics from the larger variant of the Video-FocalNet \cite{wasim2023video} family, i.e., Video-FocalNet Base (teacher) to our proposed lightweight VFL-Net (student) model, enhancing its robustness for the human action recognition task while maintaining lower computational complexity. The main contributions of this work are summarized as follows:\vspace{-0.3em}

\begin{itemize}\setlength{\itemsep}{0em}
    \item We thoroughly investigate different design configurations of Video-FocalNet family architectures and develop a computationally efficient VFL-Net model, which has only 22M parameters, optimized for spatio-temporal context modeling. The proposed VFL-Net, based on the nano spatio-temporal focal modulation design, achieves an optimal trade-off between model performance and computational efficiency.
    \item We leverage knowledge distillation to transfer the rich spatio-temporal semantics from the larger variant of the Video-FocalNet family, namely Video-FocalNet Base (teacher) having 157M parameters, to our proposed lightweight VFL-Net (student) model having 22M parameters only. This approach enhances the model's robustness for the human action recognition task while ensuring reduced computational complexity.
    \item We conduct extensive experiments on three HAR benchmarks, including UCF50, UCF101, HMDB51, SSV2, and Kinetics-400 datasets. First, we evaluate the performance of our proposed VFL-Net in comparison to the different Video-FocalNet variants, focusing on the tradeoff between model complexity and Top-1 accuracy, as depicted in Figure~\ref{fig:acc_vs_complexity}. Next, we compare the distilled version of our model (DVFL-Net) with existing state-of-the-art HAR methods. Our DVFL-Net achieves state-of-the-art Top-1 accuracy across the UCF50, UCF101, HMDB51, SSV2, and Kinetics-400 datasets while demonstrating superior computational efficiency, including reduced memory consumption and lower GFLOPs, making it a highly efficient solution for HAR tasks.
\end{itemize}
\indent\indent  The rest of this paper is organized as follows: Section II reviews related work on human action recognition. Section III explains the proposed DVFL-Net and its technical components. In Section IV, we present extensive experiments and the obtained results. Finally, Section V concludes the paper.
\section{Related Works}
Video recognition has evolved significantly with the advent of deep learning models, particularly Convolutional Neural Networks (CNNs) and Transformers, which have demonstrated remarkable capabilities in processing and understanding video data. The challenge of balancing computational speed and accuracy remains a critical focus in this field, especially as video data continues to grow in volume and complexity. Knowledge distillation techniques have emerged as a promising solution to address these challenges by enabling the transfer of knowledge from larger, more complex models (teachers) to smaller, more efficient models (students).

CNNs, particularly 3D CNNs, have been widely adopted for video recognition tasks due to their ability to capture spatiotemporal features effectively. Research indicates that 3D CNNs outperform traditional 2D CNNs by incorporating temporal information alongside spatial features, which is crucial for understanding actions in videos~\cite{anvarov2020, wangwang2021}. However, while 3D CNNs provide superior accuracy, they are computationally intensive, making them less suitable for real-time applications~\cite{lingan2019}. Similarly, multistream CNNs have demonstrated remarkable performance in the video action recognition domain by processing different streams to capture comprehensive representations of dynamic scenes. For instance, Wang et al.\cite{wang2021multi} propose Multi-Stream Interaction Networks (MSIN), which utilize dedicated streams for human skeletal dynamics, object appearance, and their interrelations to capture the nuanced interplay between humans and objects. Further, Li et al.\cite{li2023spatio} presented the Spatio-Temporal Adaptive Network (STANet), which exploits bidirectional temporal differences to adaptively fuse static semantic cues with dynamic motion information, offering an efficient alternative to traditional 3D convolutional methods. Chen et al.\cite{chen2023agpn} introduced the Action Granularity Pyramid Network (AGPN), which employs a hierarchical pyramid structure alongside multiple frame rate integration and spatio-temporal anchoring to fuse multi-granularity features, thereby enhancing recognition performance on large-scale datasets. Other works, such MAWKDN \cite{quan2023mawkdn} a multimodal fusion framework that leverages cross-view attention and wavelet-based knowledge distillation to integrate complementary video and wearable sensor data, significantly enhancing robustness under challenging conditions.

On the other hand, Transformers have gained attention in video recognition due to their self-attention mechanisms, which allow for capturing long-range dependencies in data~\cite{moutiksekkat2023}. The comparative analysis of CNNs and Transformers reveals that while CNNs excel in spatial feature extraction, Transformers are better at capturing the complex relationships within video frames, leading to improved accuracy in tasks such as action recognition~\cite{moutiksekkat2023}. However, the computational cost associated with Transformers can be significant, necessitating the exploration of more optimized and computationally efficient approaches. To achieve the best of both designs (i.e., CNN and Transformer), Waseem et al.\cite{wasim2023video} introduced a spatio-temporal focal modulation approach called Video-FocalNet, that reverses the interaction and aggregation steps of self-attention to enhance efficiency. They further implemented these steps using efficient convolution and element-wise multiplication operations, which are computationally less expensive than self-attention for video representation learning tasks.

Despite its effectiveness in capturing spatio-temporal features, Video-FocalNet faces computational efficiency limitations. Although the model replaces self-attention with convolution and element-wise multiplication to reduce complexity, Processing long video sequences or high-resolution frames remains resource-intensive. This can hinder real-time applications, especially on devices with constrained computational power. Further optimizations and model compression techniques may be required to achieve a better balance between performance and efficiency.

Knowledge distillation has been effectively utilized in video recognition to enhance model efficiency and accuracy. Traditional knowledge distillation involves training a smaller student model to replicate the outputs of a larger teacher model, thus enabling knowledge transfer in a more compact form.~\cite{zhaocui2022,camarenagonzalez2024}. Recent advancements in this area include decoupled knowledge distillation, which separates the learning of different types of knowledge, allowing for more targeted and effective training~\cite{liuwei2024}. For instance, VideoAdviser employs a multimodal transfer learning approach to distill knowledge from a complex teacher model to a simpler student model, significantly reducing computational requirements while maintaining high accuracy~\cite{wangzeng2023}.

Moreover, innovative frameworks such as generative model-based feature distillation have been proposed that leverage generative models to enhance the representation of features during the distillation process~\cite{WangZhao2024}. This approach not only improves the efficiency of the student model but also enhances its ability to generalize across various video recognition tasks. Additionally, studies have shown that knowledge distillation can accelerate model convergence and improve classification accuracy, particularly in scenarios with limited labeled data~\cite{camarenagonzalez2024,almushytili2021}.

In conclusion, integrating the computationally efficient spatio-temporal focal modulation with the knowledge distillation techniques, provides a robust framework for tackling challenges related to computational complexity and accuracy. Building on these assumptions, we aim to explore the architectural optimization of the Video-FocalNet framework alongside knowledge distillation techniques to develop a more efficient and robust approach for video action recognition.

\section{Proposed Method} 
\label{sec:proposedmethod}
This section provides an in-depth discussion of our proposed methodology, beginning with the \textit{Problem Formulation}, followed by an \textit{Spatio-Temporal Knowledge Distillation}, and concludes with the \textit{Architectural Overview} settings used in this work. 

\begin{figure*}[t]
	\centering
	\includegraphics[width=0.9\linewidth]{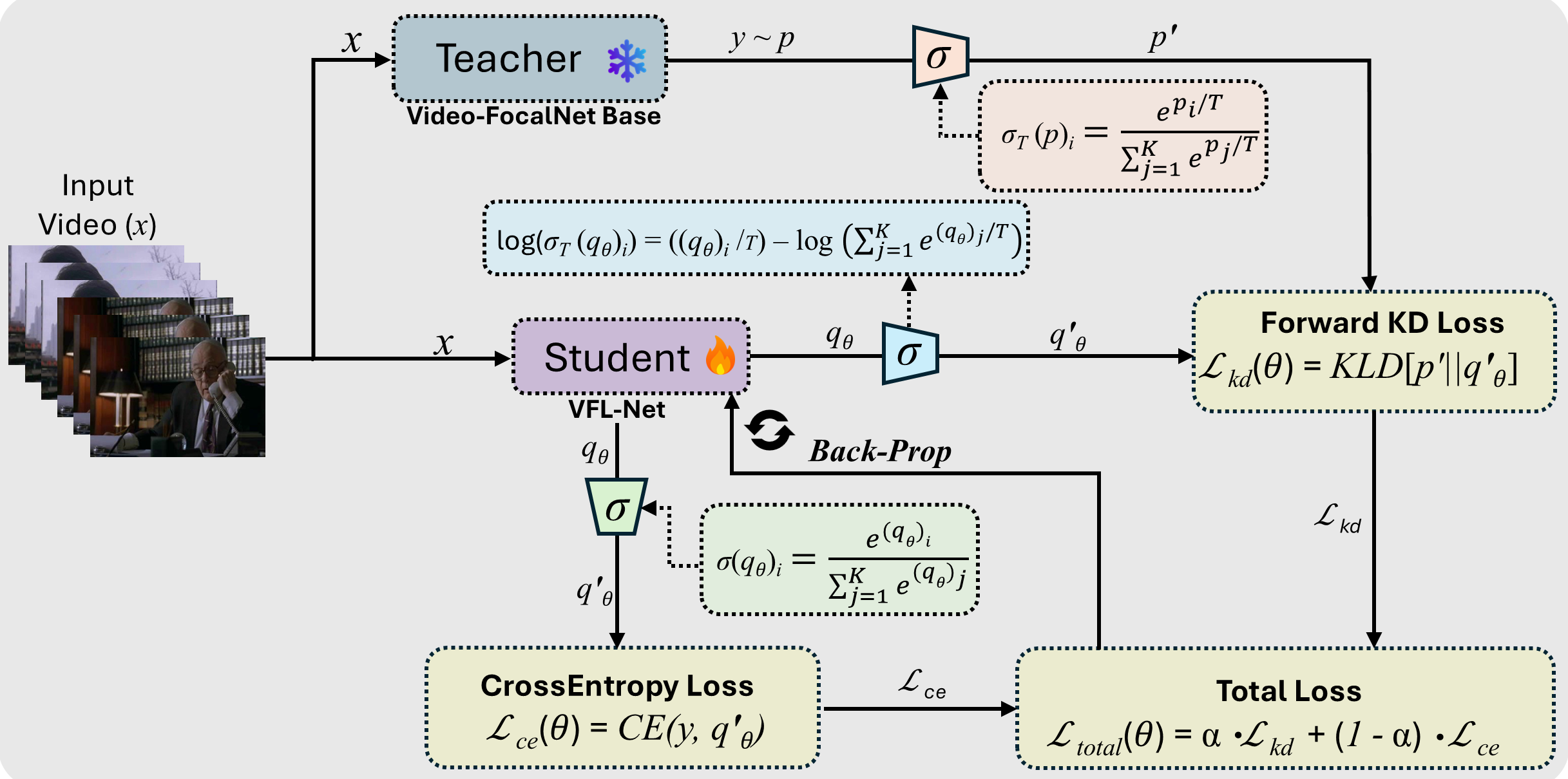}
	\caption{The figure presents the knowledge distillation framework between the teacher (Video-FocalNet Base) and student (VFL-Net) models. The input video is processed by both models, with the teacher generating soft labels \(p'\), which are compared with the student's soft labels \(q'_\theta\) to compute the \textbf{Forward KD Loss} ($\mathcal{L}_{kd}$). Simultaneously, the student is optimized using \textbf{Cross-Entropy Loss} ($\mathcal{L}_{ce}$) against the ground truth. The \textbf{Total Loss} ($\mathcal{L}_{total}$) combines knowledge distillation and cross-entropy losses, and the student model is updated via backpropagation.}
	\label{fig:kd_overview}
\end{figure*}

\subsection{Problem Formulation}
\label{sec:problem_formulation}
Human Action Recognition (HAR) aims to identify or classify a sequence of observations/actions in a given video $X = \{x_1, x_2, \dots, x_N\}$, where each $x_n \in \mathbb{R}^{H \times W \times C}$ represents a video frame at time step $n$, into a predefined set of action labels $Y = \{y_1, y_2, \dots, y_K\}$. The goal is to learn a mapping function $f : X \to Y$, where $f$ is a model capable of capturing both spatial and temporal patterns inherent in human actions.\\
\indent The core challenge in HAR lies in learning robust representations of human actions, which are inherently spatio-temporal. The temporal nature of activities necessitates modeling long-range dependencies over time, such that the model captures the evolution of temporal dynamics $T = \{t_1, t_2, \dots, t_N\}$ across frames. At the same time, each individual frame contains important spatial information, $S = \{s_1, s_2, \dots, s_N\}$, in terms of the body posture, appearance, and scene context. Therefore, HAR can be seen as optimizing the objective function:
\begin{equation}
\min_{\theta} \sum_{i=1}^{N} \mathcal{L}(f(X^{(i)}; \theta), y^{(i)}),
\label{Eq:equation_1}
\end{equation}
\indent where $\theta$ are the learnable parameters of the model, $X^{(i)}$ is the input sequence for the $i$-th sample, $y^{(i)}$ is the corresponding action label, and $\mathcal{L}$ is a loss function, typically categorical cross-entropy, that quantifies the discrepancy between the predicted and true labels.

\subsection{Spatio-Temporal Knowledge Distillation }
\label{sec:knowledge_distillation}
In this work, we focus on distilling spatio-temporal knowledge from a large pre-trained teacher model Video-FocalNet Base \cite{wasim2023video} with 157 million parameters, to a much smaller student model VFL-Net (ours) having only 22 million parameters. Knowledge distillation has emerged as a popular technique to compress deep learning models, enabling smaller models to learn from larger more powerful models without substantial loss in performance. Our goal is to reduce the computational complexity of video action recognition tasks by distilling the knowledge embedded in the teacher’s spatio-temporal representations to the student model, which makes it efficient enough for deployment on resource-constrained devices without a significant drop in accuracy. In particular, we aim to distill rich spatio-temporal information, making the student model competitive despite its smaller parameter count. The detailed visual overview of our proposed method is depicted in Figure~\ref{fig:kd_overview}.\\
\indent The spatio-temporal knowledge transfer is particularly important in our case, as we are working with video data, where both spatial and temporal dimensions are crucial for accurate action recognition. To handle the complexity of spatio-temporal information, both the teacher and student models leverage the Focal Modulation mechanism introduced in Video-FocalNet \cite{wasim2023video}. This mechanism uses a unique combination of temporal hierarchical contextualization and spatial modulation to capture long-range dependencies in both spatial and temporal dimensions. The teacher model, with its larger capacity, can model these relationships more extensively, while the student model learns to replicate these high-level abstractions under the supervision of teacher model.\\
\indent Our proposed approach follows offline knowledge distillation scheme (with response-based knowledge distillation settings), where the student model mimics the soft predictions of the pre-trained teacher model rather than just the hard labels from the ground truth. This method involves minimizing the divergence between the soft output probabilities (logits) of the teacher and student models. Formally, the teacher model produces a probability distribution over class predictions, denoted by $\mathcal{p}$, while the student model generates its own distribution, $\mathcal{q_{\theta}}$. The student is trained to minimize the difference between its predicted logits and those of the teacher, typically using Kullback-Leibler Divergence ($\mathcal{KLD}$). The distillation loss $\mathcal{L}_{kd}$ can be expressed as:
\begin{equation}
\mathcal{L}_{kd} = \mathcal{KLD}(\overbrace{\sigma(p / \tau)}^{p'}, \overbrace{\sigma(q_{\theta} / \tau)}^{q'_{\theta}}) \cdot \tau^2,
\label{Eq:equation_2}
\end{equation}
\indent Where $\mathcal{\sigma}$ represents the softmax function and $\mathcal{\tau}$ is a temperature hyper-parameter that controls the smoothness of the output probability distributions. A higher value of $\mathcal{\tau}$ leads to softer probabilities, which makes it easier for the student to learn from the teacher’s output. $\mathcal{KLD}$ estimates the divergence between two given probability distributions that include soft predictions of teacher and student models. Mathematically, $\mathcal{KLD}$ divergence from distribution $q'_{\theta}$ to distribution $p$ can be expressed as follows:
\begin{equation}
\mathcal{KLD}(p'||q'_{\theta}) = \mathbb{E}_{x \sim p'} \left[ \log \left( \frac{p'}{q'_{\theta}} \right) \right],
\label{Eq:equation_3}
\end{equation}
\begin{figure*}[t]
	\centering
	\includegraphics[width=0.9\linewidth]{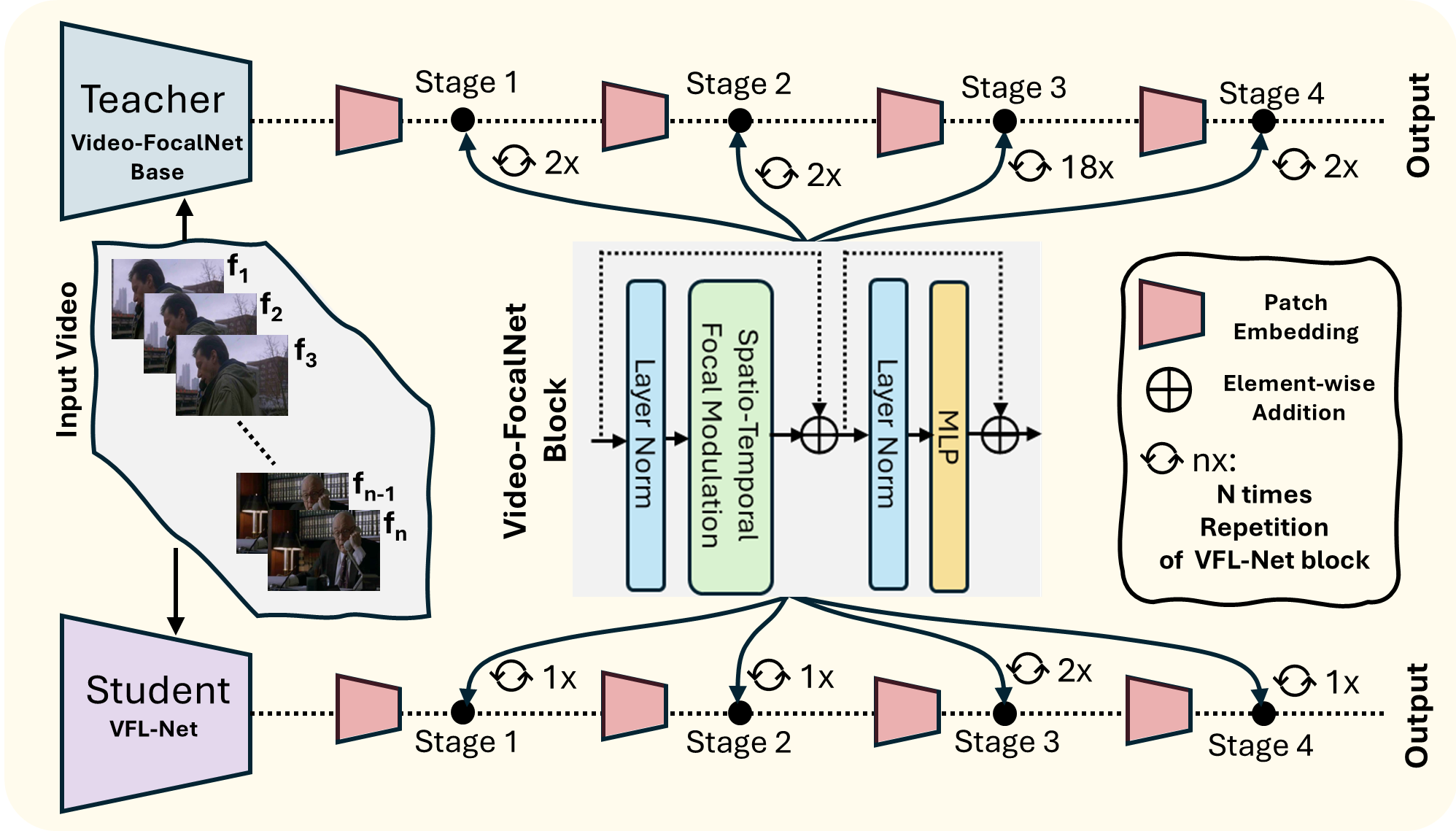}
	\caption{\textbf{The architectural overview of teacher and student models.} Both the teacher and student models follows a four-stage architecture, with each stage containing a patch embedding layer and Video-FocalNet blocks. The \textbf{teacher model} consists of 24 Video-FocalNet blocks, distributed as 2, 2, 18, and 2 blocks across stages one, two, three, and four, respectively. In contrast, the \textbf{student model} is more lightweight, containing only 5 Video-FocalNet blocks, with 1, 1, 2, and 1 blocks assigned to stages one through four, respectively.}
	\label{fig:arch_overview}
\end{figure*}
\begin{figure}[t]
	\centering
	\includegraphics[width=\linewidth]{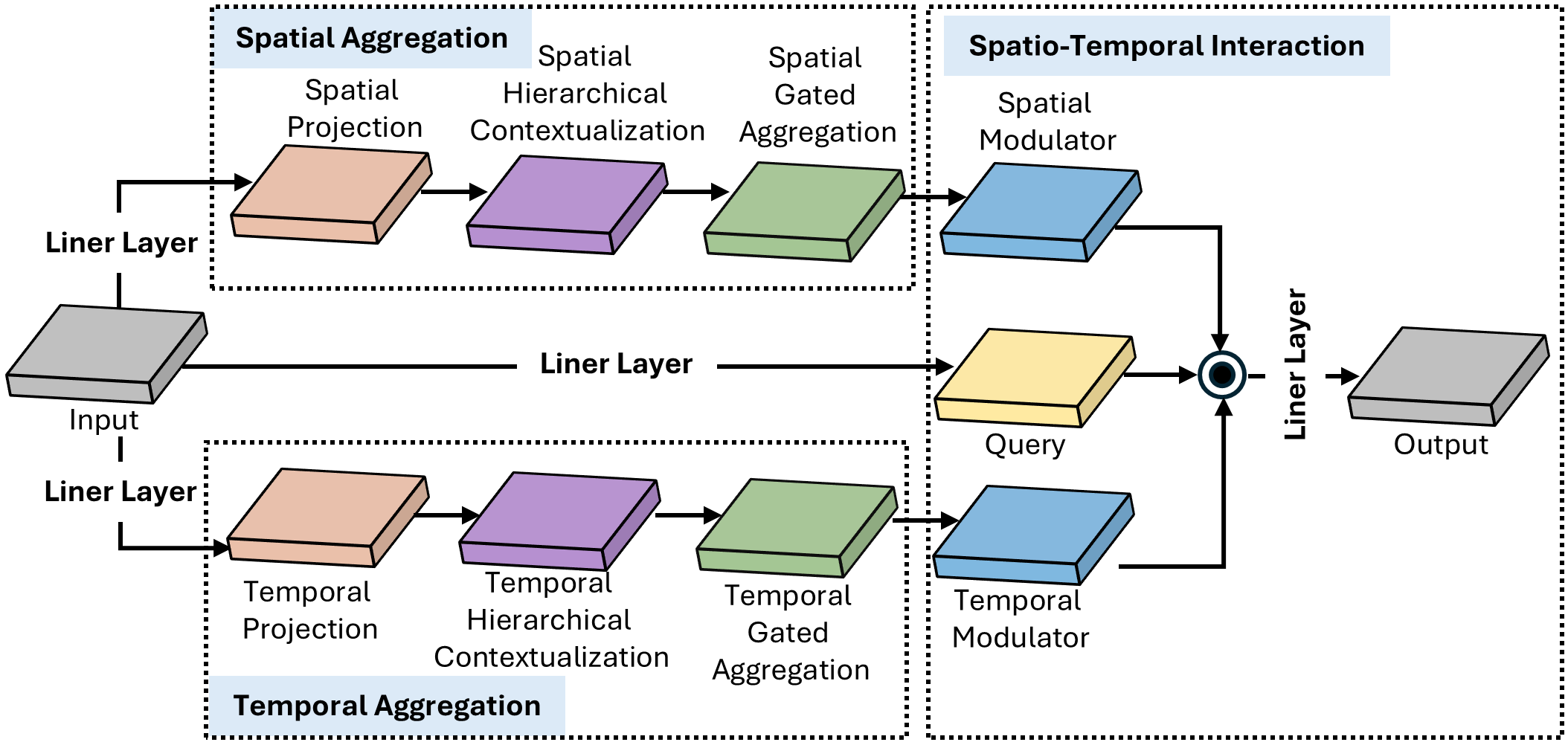}
	\caption{The \textbf{Spatio-Temporal Focal Modulation} layer, processing \textbf{spatial} and \textbf{temporal} information independently. It separates spatial and temporal encoding, allowing each to be modeled distinctly before being combined for enhanced spatio-temporal representation.}
	\label{fig:focal_module}
\end{figure}
\indent The term $\mathbb{E}_{x \sim p'}$ refers to the expected value of logarithmic function with respect to the probability distribution $p'$. Expending Eq.~\ref{Eq:equation_3} results in:
\begin{equation}
\mathcal{KLD}(p'||q'_{\theta}) = \sum_{i} p'(i) \log \left( \frac{p'(i)}{q'_{\theta}(i)} \right),
\label{Eq:equation_4}
\end{equation}
\indent Here $p'(i)$ and $q'_{\theta}(i)$ are the $i^{th}$ samples of teacher's and student's probability distributions, respectively. Whereas, $\log \left( \frac{p'(i)}{q'_{\theta}(i)} \right)$ represents the information gain between two probability distributions (i.e., $p'$ and $q'_{\theta}$) for each sample $i$. In the context of knowledge distillation, minimizing $\mathcal{KLD}$ ensures that the student model's output closely follows the teacher model's soft predictions. The standard Cross-Entropy ($\mathcal{CE}$) loss is used as a supervised criterion for the student model to learn from ground truth labels in supervised settings: 
\begin{equation}
\mathcal{L}_{ce} = CE(y, \overbrace{\sigma(q_{\theta})}^{q'_{\theta}})),
\label{Eq:equation_5}
\end{equation}
\indent Here $y$ is the ground truth label and $q'_{\theta}$ is the output probability distribution of student model:
\begin{equation}
\mathcal{L}_{total} = \alpha \cdot \mathcal{L}_{kd} + (1 - \alpha) \cdot \mathcal{L}_{ce}, 
\label{Eq:equation_6}
\end{equation}
\indent Here $\mathcal{L}_{total}$ is the weighted sum of two losses that included $\mathcal{L}_{kd}$ and $\mathcal{L}_{ce}$, where $\alpha$ is hyper-parameter that balances the contributions of the distillation loss $\mathcal{L}_{kd}$ and student loss $\mathcal{L}_{ce}$. This formulation allows the student model to benefit both from the direct supervision of the ground truth labels and the more refined knowledge of the teacher’s output.   

\subsection{Architectural Overview}
\label{sec:architectural_overview}
\subsubsection{Spatio-Temporal Focal Modulation}
The Spatio-Temporal Focal Modulation mechanism extends focal modulation to capture both spatial and temporal dimensions in videos. Given a spatio-temporal feature map $X_{st} \in \mathbf{R}^{T \times H \times W \times C}$, where T is the number of frames, $H \times W$ is the spatial resolution, and C is the number of channels, the model independently aggregates spatial and temporal context. A spatial slice $X_{st,t} \in \mathbf{R}^{H \times W \times C}$ captures context at each time step t, while a temporal slice $X_{st,hw} \in \mathbf{R}^{T \times C}$ captures temporal context across frames at spatial position (h,w). The interaction of the spatial and temporal modulators with the query token is expressed by:
\begin{equation}
y_{i} = T_{st} (M_{s}(i_{t}, x_{st,t}), M_{t}(i_{hw}, x_{st,hw}), x_{i}), 
\label{Eq:equation_7}
\end{equation}
\indent where $M_{s}$ and $M_{t}$ are the spatial and temporal modulators, respectively, and $T_{st}$ is the interaction function. The spatio-temporal focal modulation process includes \textbf{Hierarchical Contextualization} and \textbf{Gated Aggregation}, where the spatial and temporal modulators are learned through depthwise and pointwise convolutions. These modulators interact with the query tokens via element-wise multiplication. This is described as follows: 
\begin{equation}
y_i = q(x_i) \odot m_s(i_t, X_{st,t}) \odot m_t(i_{hw}, X_{st,hw}), 
\label{Eq:equation_8}
\end{equation}
\indent Here, $q(x_{i})$ is the query projection, and $m_{s}$ and $m_{t}$ represent the spatial and temporal aggregation functions. This formulation efficiently captures local and global spatio-temporal dependencies, enabling rich feature representations with reduced computational complexity. The detailed architectural overview diagram of teacher and student models are depicted in Figure~\ref{fig:arch_overview}.
\subsubsection{Spatio-Temporal Hierarchical Contextualization}
The input spatio-temporal feature map $X_{st} \in \mathbf{R}^{T \times H \times W \times C}$ is first projected into spatial and temporal components using two linear layers to generate $Z_s^0$ and $Z_t^0$, as expressed follows:
\begin{equation}
\begin{aligned}
    Z_s^0 = f_{z,s}(X_{st}) \in \mathbf{R}^{T \times H \times W \times C}, \\
    Z_t^0 = f_{z,t}(X_{st}) \in \mathbf{R}^{T \times H \times W \times C},
\end{aligned}
\label{Eq:equation_9}
\end{equation}
\indent Here, $f_{z,s}$ and $f_{z,t}$ are spatial and temporal linear projection functions, respectively. These initial projections are passed through a series of depthwise and pointwise convolutions to progressively capture spatial and temporal contextual information. At each focal level $\ell$, the outputs $Z_s^\ell$ and $Z_t^\ell$ are generated by applying the GeLU activation function, as follows:
\begin{equation}
\begin{aligned}
    Z_s^\ell = f_a^\ell,s(Z_s^{\ell-1}) \overset{\Delta}{=} \text{GeLU}(\text{DWConv}(Z_s^{\ell-1})) \in \mathbb{R}^{T \times H \times W \times C}, \\
    Z_t^\ell = f_a^\ell,t(Z_t^{\ell-1}) \overset{\Delta}{=} \text{GeLU}(\text{PWConv}(Z_t^{\ell-1})) \in \mathbb{R}^{T \times H \times W \times C},
\end{aligned}
\label{Eq:equation_10}
\end{equation}
\indent The terms $f_{a,s}^{l}(\cdot)$ and $f_{a,t}^{l}(\cdot)$ represent the spatial and temporal contextualization functions, respectively, each utilizing the GeLU \cite{hendrycks2016gaussian} activation function. At the final level $L$, a global average pooling operation is performed along the spatial and temporal dimensions to produce the global representations $Z_s^{L+1}$ and $Z_t^{L+1}$, as expressed follows:
\begin{equation}
\begin{aligned}
    Z_s^{L+1} = \text{AvgPool}(Z_s^L), \\
    \quad Z_t^{L+1} = \text{AvgPool}(Z_t^L),
\end{aligned}
\label{Eq:equation_11}
\end{equation}
\indent Here, $Z_s^L$ and $Z_t^L$ represent spatial and temporal features at level $L$, respectively. Whereas $AvgPool$ is global average pooling.
\subsubsection{Spatio-Temporal Gated Aggregation}
Next, the spatial and temporal feature maps are condensed into modulators through a gating mechanism. The spatial and temporal gating weights $G_s = f_{g,s}(X_{st}) \in R^{H \times W \times (L+1)}$ and $G_t = f_{g,t}(X_{st}) \in R^{T \times (L+1)}$ are derived using linear projection layers $f_{g,s}$ and $f_{g,t}$, respectively. Then, a dot product is performed between the feature maps and their corresponding gates as follows:
\begin{equation}
\small
\begin{aligned}
    Z^{out}_s & = \sum_{\ell=1}^{L+1} G_s^\ell \odot Z_s^\ell \in R^{H \times W \times C},\\
    Z^{out}_t & = \sum_{\ell=1}^{L+1} G_t^\ell \odot Z_t^\ell \in R^{T \times C},
\end{aligned}
\label{Eq:equation_11}
\end{equation}
\indent Where $Z_s^\ell$ and $Z_t^\ell$ represent the spatial and temporal feature maps, and $G_s^\ell$ and $G_t^\ell$ are the gating values at each level $\ell$. These outputs are used to generate the spatial and temporal modulators $M_s = h_s(Z_{\text{out}}^s)$ and $M_t = h_t(Z_{\text{out}}^t)$. Finally, the spatio-temporal focal modulation is applied as follows:
\begin{equation}
    y_i = q(x_i) \odot h_s(\sum_{\ell=1}^{L+1} g^\ell_{i,s} \cdot z^\ell_{i,s}) \odot h_t(\sum_{\ell=1}^{L+1} g^\ell_{i,t} \cdot z^\ell_{i,t})
    \label{Eq:equation_12}
\end{equation}
\indent Here, $z_{i,s}^\ell$ and $z_{i,t}^\ell$ represent the spatial and temporal visual features at location $i$ in the feature maps $Z^\ell_s$ and $Z^\ell_t$, respectively. Similarly, $g_{i,s}^\ell$ and $g_{i,t}^\ell$ correspond to the spatial and temporal gating values at the same location in the gating maps $G^\ell_s$ and $G^\ell_t$. These gating values modulate the feature maps to capture relevant spatio-temporal dependencies efficiently. For better understanding, the detailed visual overview of spatio-temporal focal modulation is depicted in Figure~\ref{fig:focal_module}.
\begin{table}[t]
\caption{Architectural configurations of teacher and student models.}
\centering
\begin{tabular}{c|c|c}
\hline
 Configuration & Teacher & Student \\
\hline
Architecture Type &  Video-FocalNet Base \cite{wasim2023video} & VFL-Net  \\
Embedding Dim & 128 & 96 \\
Depths & [2,2,18,2]& [1,1,2,1] \\
Focal Levels & [2,2,2,2] & [2,2,2,2]  \\
Focal Windows & [3,3,3,3] & [3,3,3,3]  \\
Drop Path Rate & 0.5 & 0.2  \\
\hline
\end{tabular}
\label{tab:network_config}
\end{table}%
\begin{table}[t]
\caption{Training (both pretraining and knowledge distillation) hyperparameters used in the experiments of this work.}
\centering
\begin{tabular}{l l}
\toprule
Hyperparameter & Value \\
\hline
\rowcolor{gray!10}\textit{\textbf{Training}} &\\
Optimizer &  SGD  \\
Batch size &  8 \\
Epochs & 120 \\
Number of frames & 8 \\
Learning rate schedule & cosine with linear warmup \\
Warmup epochs & 20  \\
Base learning rate & 0.1 \\
Warmup learning rate & 0.001 \\
\hline
\rowcolor{gray!10}\textit{\textbf{Knowledge Distillation}} & \\
Alpha ($\alpha$) & \{0.3, 0.5, 0.7\}\\
Temperature ($\tau$) & \{5, 10, 15\}\\
\bottomrule
\end{tabular}
\label{tab:huper-parameters}
\end{table}%
\subsection{Networks Design}
\subsubsection{Teacher}
The teacher model, Video-FocalNet Base is used from \cite{wasim2023video}, focusing on enhancing the spatio-temporal representations, a significant element for accurate video action recognition tasks. Video-FocalNet Base employs a unique focal modulation mechanism, allowing the model to capture long-range dependencies while keeping the computational complexity manageable. By focusing on both local and global features within video frames, the model achieves a fine balance between accuracy and efficiency. The network configurations of the teacher model is presented in Table~\ref{tab:network_config}.\\
\indent The teacher model operates at a larger scale, with higher computational capacity due to its increased number of parameters and layers. This allows Video-FocalNet Base to process richer spatio-temporal details from video sequences, leading to superior action recognition performance. Specifically, its deep architecture leverages multiple focal blocks that process different receptive fields, combining fine-grained details with long-range context, essential for understanding complex motion dynamics in videos. These architectural strengths enable the teacher model to achieve high accuracy, which is critical when used as a supervisory guide for training smaller student models through distillation techniques.

\subsubsection{Student}
The student model, VFL-Net is the nano variant of Video-FocalNet Base, introduced in this paper as a more lightweight version of the teacher model. It inherits the core focal modulation principles but with a significant reduction in parameters and computational complexity. The student model is designed to maintain a competitive level of accuracy while being more resource-efficient, making it suitable for deployment in scenarios where hardware limitations are a concern, such as mobile devices or edge computing platforms. Despite its compact design, the VFL-Net is designed to distill essential spatio-temporal knowledge from the teacher model, ensuring that the critical action recognition capabilities are preserved even with fewer computational resources.\\
\indent Architecturally, the student model's design reduces the number of focal blocks and layers compared to its teacher counterpart while carefully selecting the most essential components to retain performance as given in Table~\ref{tab:network_config}. The distillation process plays a pivotal role in transferring knowledge from the teacher model, allowing the student model to achieve competitive results despite its smaller size. By leveraging the strengths of the teacher model’s focal modulation mechanism, the Nano variant strikes an optimal balance between accuracy and efficiency, making it a powerful tool for video action recognition in resource-constrained environments.

\section{Experimental Results and Discussion} \label{sec:experimentalresults}
\subsection{Datasets}
We conducted extensive experiments on three publicly available action recognition datasets, including UCF50\cite{reddy2013recognizing}, UCF101\cite{soomro2012ucf101}, and HMDB51\cite{kuehne2011hmdb} datasets. The UCF50 dataset contains 6,618 videos across 50 action categories, split into five groups for training and validation, with 70\% used for training and 30\% for validation. The UCF101 dataset extends this with 101 action categories and 13,320 videos, divided into three standard training and test splits. Similarly, HMDB51 consists of 6,849 video clips across 51 action categories, with a 70\%-30\% training-validation split. Both UCF101 and HMDB51 present significant challenges due to large intra-class variations, complex backgrounds, and varying camera angles, making them ideal for evaluating the VFL-Net model. The SSV2 \cite{goyal2017something} dataset is a collection of 220,847 short video clips across 174 fine-grained action classes, focusing on human-object interactions with subtle motion differences. It’s widely used for benchmarking models in video action recognition tasks. The Kinetics-400 \cite{kay2017kinetics} dataset contains 306,245 video clips spanning over 400 human action classes, sourced from YouTube. It provides diverse, large-scale data for training and evaluating action recognition models in complex, real-world scenarios.

\subsection{Evaluation Metrics}
In this work, we evaluate the performance of our method using top-1 and top-5 accuracy metrics. For baseline results on UCF50, UCF101, HMDB51, SSV2, and Kinetics-400, we report both top-1 and top-5 accuracy. However, for comparison with state-of-the-art methods, we primarily focus on presenting the top-1 accuracy.

\subsection{Implementation Details}
Our training protocol follows the scheme used in ~\cite{wasim2023video, li2022uniformer}, using $120$ epochs with a linear warmup over the first $20$ epochs and the SGD optimizer. The learning rate is linearly scaled by $LR \times \frac{batchsize}{512}$, where $LR = 0.1$ serves as the base learning rate. Following ~\cite{wasim2023video}, we sample $T$ frames (8 frames in our case) during training with a stride of $\tau$, denoted as $T \times \tau$. For the spatial domain, as in \cite{wasim2023video,szegedy2015inception}, we apply a crop of $H \times W = 224 \times 224$,, with the input area selected from a scale range of $[\min, \max] = [0.08, 1.00]$. While inferencing, we report results as an average across $N_{clip} \times N_{crops}$, where $N_{clip}$  clips are uniformly sampled from the video, and $N_{crops}$  spatial crops are taken per clip. In our case, we evaluate our method using both single-view inference (1 clip and 1 crop) and 12-view inference (4 clips and 3 crops). The hyperparameters used in knowledge distillation algorithm including alpha ($\alpha$) and temperature ($\tau$) are assigned with different values to see their impacts on the overall performance. The hyperparameters used for training and knowledge distillation, along with their corresponding values, are provided in Table~\ref{tab:huper-parameters}. \\
\indent All experiments are conducted using CUDA (version 12.4) enabled PyTorch 2.4 on a Linux-based system (Ubuntu) equipped with an AMD Ryzen Threadripper PRO processor (32 cores), 512GB RAM, and three NVIDIA RTX 6000 Ada GPUs (48GB each).

\subsubsection{Teacher's Initialization}
The teacher model (i.e., Video-FocalNet Base) is first pretrained on a dataset from the datasets used in this work (i.e., UCF50, UCF101, HMDB51, SSV2, and Kinetics-400). During pretraining, we initialize the teacher model with the ImageNet-1K pretrained weights.
\subsubsection{Student's Initialization}
In knowledge distillation training, the student model (i.e., VFL-Net) is trained from the scratch under the supervision of the teacher model (pretrained on the same dataset). 

\subsection{Quantitative Evaluation of Baseline Models}
This section quantitatively evaluates and compares our proposed method, VFL-Net, with baseline models on the UCF50, UCF101, HMDB51, SSV2, and Kinetics-400 datasets. Results are reported in Tables~\ref{tab:baseline_ucf50}, \ref{tab:baseline_ucf101}, \ref{tab:baseline_hmdb51}, \ref{tab:baseline_ssv2}, and \ref{tab:baseline_k400} respectively. Table~\ref{tab:baseline_ucf50} demonstrates that VFL-Net achieves a significantly better tradeoff between model size and performance on the UCF50 dataset compared to the baseline models. While Video-FocalNet Base, Small, and Tiny models achieve slightly higher Top-1 and Top-5 accuracy (91.1\%, 90.8\%, and 90.1\% in Top-1), they come at the cost of much larger model sizes of 157M, 88M, and 49M parameters, respectively. In contrast, VFL-Net utilizes only 22M parameters, which is over 7$\times$ smaller than Video-FocalNet Base and more than 2$\times$ smaller than Video-FocalNet Tiny, yet still maintains a reasonable Top-1 accuracy of 81.4\% and Top-5 accuracy of 91.4\%.\\
\indent Similarly, Table~\ref{tab:baseline_ucf101} demonstrates the efficacy of our VFL-Net on the UCF101 dataset by achieving a better tradeoff between the model's parameter size and accuracy (Top-1 and Top-5). Despite having significantly fewer parameters, VFL-Net achieves a Top-1 accuracy of 82.5\% and a Top-5 accuracy of 93.7\%. In contrast, the larger Video-FocalNet models, while attaining slightly higher accuracy (91.9\% Top-1 for the Video-FocalNet Base model), require substantially more parameters of 157M for Base, 88M for Small, and 49M for Tiny. Thus, VFL-Net offers reasonable performance, with over 7$\times$ fewer parameters than Video-FocalNet Base.\\
\indent Table \ref{tab:baseline_hmdb51} provides further evidence of VFL-Net's strong balance between accuracy and model efficiency on the HMDB51 dataset. Although the Video-FocalNet Base, Small, and Tiny models achieve higher Top-1 accuracy (84.2\%, 83.4\%, and 81.5\%, respectively), this comes with a significant increase in parameters of 157M for Base, 88M for Small, and 49M for Tiny. In comparison, VFL-Net achieves a reasonable Top-1 accuracy of 71.6\% and Top-5 accuracy of 88.5\% with only 22M parameters. When evaluating on SSV2 dataset (Table \ref{tab:baseline_ssv2}), our VFL-Net achieves 64.7\% Top-1 and 72.4\% Top-5 accuracy, demonstrating competitive performance despite its significantly smaller size compared to the baseline models. In contrast, Video-FocalNet Tiny, Small, and Base models, which have 49M, 88M, and 157M parameters, respectively, achieve Top-1 accuracies of 67.4\%, 69.8\%, and 71.0\%, along with higher Top-5 accuracies. Despite the smaller parameter count (having only 22M parameters), our VFL-Net maintains a strong performance, showing a decent tradeoff between model complexity and accuracy.\\
\begin{table}[t]
\caption{\small Performance comparison of our VFL-Net with baseline models on the \textbf{UCF50} dataset, highlighting Top-1 and Top-5 accuracy, along with the number of parameters of each model.}
\centering
\scriptsize
\begin{tabular}{l|c|c|c|c}
\hline
\rowcolor{white!10}Model & Pretrain & Top-1 & Top-5 & Parameters\\
\hline
\rowcolor{gray!10}Video-FocalNet Tiny \cite{wasim2023video} & ImageNet-1K &  90.1 & 97.1 & 49M  \\ 
\rowcolor{white!10}Video-FocalNet Small \cite{wasim2023video} & ImageNet-1K &  90.8 & 97.9 & 88M   \\
\rowcolor{gray!10}Video-FocalNet Base \cite{wasim2023video} & ImageNet-1K & 91.1 & 98.7 & 157M   \\
\rowcolor{blue!10}VFL-Net (\textbf{Ours}) & ImageNet-1K &  81.4 & 91.4 & 22M  \\ 
\hline
\end{tabular}
\label{tab:baseline_ucf50}
\end{table}
\begin{table}[t]
\caption{\small Performance comparison of our VFL-Net with baseline models on the \textbf{UCF101} dataset, highlighting Top-1 and Top-5 accuracy, along with the number of parameters of each model.}
\centering
\scriptsize
\begin{tabular}{l|c|c|c|c}
\hline
\rowcolor{white!10}Model & Pretrain & Top-1 & Top-5 & Parameters \\
\hline
\rowcolor{gray!10}Video-FocalNet Tiny \cite{wasim2023video} & ImageNet-1K & 90.7 & 98.4 & 49M  \\
\rowcolor{white!10}Video-FocalNet Small \cite{wasim2023video} & ImageNet-1K & 91.2 & 98.9 & 88M   \\
\rowcolor{gray!10}Video-FocalNet Base \cite{wasim2023video} & ImageNet-1K & 91.9 & 99.4 & 157M   \\
\rowcolor{blue!10}VFL-Net (\textbf{Ours}) & ImageNet-1K & 82.5 & 93.7 & 22M  \\ 
\hline
\end{tabular}
\label{tab:baseline_ucf101}
\end{table}
\begin{table}[t]
\caption{\small Performance comparison of our VFL-Net with baseline models on the \textbf{HMDB51} dataset, highlighting Top-1 and Top-5 accuracy, along with the number of parameters of each model.}
\centering
\scriptsize
\begin{tabular}{l|c|c|c|c}
\hline
\rowcolor{white!10}Model & Pretrain & Top-1 & Top-5 & Parameters \\
\hline
\rowcolor{gray!10}Video-FocalNet Tiny \cite{wasim2023video} & ImageNet-1K & 81.5 & 91.1 & 49M  \\ 
\rowcolor{white!10}Video-FocalNet Small \cite{wasim2023video} & ImageNet-1K &83.4 & 92.4 & 88M   \\
\rowcolor{gray!10}Video-FocalNet Base \cite{wasim2023video} & ImageNet-1K & 84.2 & 93.6 & 157M   \\
\rowcolor{blue!10}VFL-Net (\textbf{Ours}) & ImageNet-1K & 71.6 & 88.5 & 22M  \\ 
\hline
\end{tabular}
\label{tab:baseline_hmdb51}
\end{table}
\begin{table}[t]
\caption{\small Performance comparison of our VFL-Net with baseline models on the \textbf{SSV2} dataset, highlighting Top-1 and Top-5 accuracy, along with the number of parameters of each model.}
\centering
\scriptsize
\begin{tabular}{l|c|c|c|c}
\hline
\rowcolor{white!10}Model & Pretrain & Top-1 & Top-5 & Parameters \\
\hline
\rowcolor{gray!10}\textcolor{red}{$\dagger$}Video-FocalNet Tiny \cite{wasim2023video}  & ImageNet-1K & 67.4 & 76.5 & 49M  \\ 
\rowcolor{white!10}\textcolor{red}{$\dagger$}Video-FocalNet Small \cite{wasim2023video}  & ImageNet-1K & 69.8 & 78.1 & 88M   \\
\rowcolor{gray!10}\textcolor{red}{$\dagger$}Video-FocalNet Base \cite{wasim2023video}  & ImageNet-1K & 71.0 & 79.3 & 157M   \\
\rowcolor{blue!10}VFL-Net (\textbf{Ours}) & ImageNet-1K & 64.7 & 72.4 & 22M  \\ 
\hline
\end{tabular}
\label{tab:baseline_ssv2}
\caption*{\footnotesize Note: The symbol \textcolor{red}{$\dagger$} indicates that the results are reproduced locally and not directly reported from Video-FocalNet \cite{wasim2023video}.}
\end{table}
\begin{table}[h!]
\caption{\small Performance comparison of our VFL-Net with baseline models on the \textbf{Kinetics-400} dataset, highlighting Top-1 and Top-5 accuracy, along with the number of parameters of each model.}
\centering
\scriptsize
\begin{tabular}{l|c|c|c|c}
\hline
\rowcolor{white!10}Model & Pretrain & Top-1 & Top-5 & Parameters \\
\hline
\rowcolor{gray!10}\textcolor{red}{$\dagger$}Video-FocalNet Tiny \cite{wasim2023video} & ImageNet-1K & 79.6 & 87.8 & 49M  \\
\rowcolor{white!10}\textcolor{red}{$\dagger$}Video-FocalNet Small \cite{wasim2023video} & ImageNet-1K & 81.3 & 89.6 & 88M   \\
\rowcolor{gray!10}\textcolor{red}{$\dagger$}Video-FocalNet Base \cite{wasim2023video}  & ImageNet-1K & 83.1 & 90.4 & 157M   \\
\rowcolor{blue!10}VFL-Net (\textbf{Ours}) & ImageNet-1K & 77.5 & 85.9 & 22M  \\ 
\hline
\end{tabular}
\label{tab:baseline_k400}
\caption*{\footnotesize Note: The symbol \textcolor{red}{$\dagger$} indicates that the results are reproduced locally and not directly reported from Video-FocalNet \cite{wasim2023video}.}
\end{table}
\begin{table}[t]
\caption{\small Performance improvements of our DVFL-Net with knowledge distillation on \textbf{UCF50}, \textbf{UCF101}, \textbf{HMDB51}, \textbf{SSV2}, and \textbf{Kinetics-400} datasets, showing notable gains (\textcolor{green}{denoted by $\uparrow$}) in both Top-1 and Top-5 accuracy.}
\centering
\scriptsize
\begin{tabular}{l|c|c|c|c}
\hline
\rowcolor{white!10}Model & Dataset & Pretrain &Top-1 & Top-5 \\
\hline
\rowcolor{gray!10}DVFL-Net & UCF50 & ImageNet-1K & 86.6 (\textcolor{green}{$\uparrow$ 5.2}) & 95.2 (\textcolor{green}{$\uparrow$ 3.8})    \\
\rowcolor{white!10}DVFL-Net & UCF101 & ImageNet-1K & 88.4 (\textcolor{green}{$\uparrow$ 5.9}) & 96.1 (\textcolor{green}{$\uparrow$ 2.4})     \\
\rowcolor{gray!10}DVFL-Net & HMDB51 & ImageNet-1K & 82.7 (\textcolor{green}{$\uparrow$ 11.1}) & 92.8 (\textcolor{green}{$\uparrow$ 4.3})  \\ 
\rowcolor{white!10}DVFL-Net & SSV2 & ImageNet-1K & 70.8 (\textcolor{green}{$\uparrow$ 6.1}) & 78.7 (\textcolor{green}{$\uparrow$ 6.3}) \\ 
\rowcolor{gray!10}DVFL-Net & Kinetics-400 & ImageNet-1K & 83.1 (\textcolor{green}{$\uparrow$ 5.6}) & {89.3 (\textcolor{green}{$\uparrow$ 3.9})} \\ 
\hline
\end{tabular}
\label{tab:kd_results}
\end{table}
\begin{table}[t]
\caption{\small Comparison of DVFL-Net with state-of-the-art human action recognition methods on the \textbf{UCF50} dataset. Top-1 accuracy is used as the evaluation metric for comparison, where '---' indicates the unavailable information.}
\centering
\begin{tabular}{l|c|c|c|c}
\hline
\rowcolor{white!10}Method & Venue & Frames & Parameters &Top-1  \\
\hline
\rowcolor{gray!10}LGF+QSVM \cite{al2021making} & ACCESS'21 & --- & --- & 69.4\\
\rowcolor{white!10}DTEF \cite{ramya2021human} & MMTA'21 & 30 & --- & 80.0\\
\rowcolor{gray!10}HDENN \cite{dasari2022human} & IJCNN'22 & 20 & 6.6M & 75.0\\
\rowcolor{white!10}SAG-DLM \cite{liu2022simple} & NURCMPT'22 & --- & --- & 82.1\\
\rowcolor{gray!10}3D-CNN \cite{vrskova2022human} & APPLSCI'22 & --- & --- & 82.6\\
\rowcolor{white!10}ACE-BET \cite{liu2022fast} & ACCESS'22 & --- & --- & 84.0\\
\rowcolor{gray!10}CLST-DNN \cite{saif2023spatio} & IJACSA'23 & 10 & --- & 80.8\\
\rowcolor{white!10}AFC-BiGRU \cite{chopra2023human} & IEEESMC'23 & --- & --- & 85.8\\
\rowcolor{blue!10}DVFL-Net (\textbf{Ours}) &  & 8 & 22M & \textbf{86.6}   \\
\hline
\end{tabular}
\label{tab:sota_ucf50}
\end{table}
\begin{table}[h!]
\caption{\small Comparison of DVFL-Net with state-of-the-art human action recognition methods on the \textbf{UCF101} dataset. Top-1 accuracy is used as the evaluation metric for comparison, where '---' indicates the unavailable information.}
\centering
\begin{tabular}{l|c|c|c|c}
\hline
\rowcolor{white!10}Method & Venue & Frames & Parameters &Top-1  \\
\hline
\rowcolor{gray!10}STS \cite{wang2021self} & TPAMI'21 & 16 & 9.6M & 77.8\\
\rowcolor{white!10}MFO \cite{qian2021enhancing} & ICCV'21 & 16 & --- & 79.1\\
\rowcolor{gray!10}VideoMoCo \cite{pan2021videomoco} & CVPR'21 & 16 & 14.4M & 78.7\\
\rowcolor{white!10}SVC2D \cite{kumawat2022action} & TPAMI'22 & 16 & 60.5M & 58.1\\
\rowcolor{gray!10}TCLR \cite{dave2022tclr} & CVIU'22 & 16 & --- & 85.4\\
\rowcolor{white!10}L2A \cite{gowda2022learn2augment} & ECCV'22 & 16 & --- & 73.1\\
\rowcolor{gray!10}LTG \cite{xiao2022learning} & CVPR'22 & 8 & --- & 79.3\\
\rowcolor{white!10}PPTK \cite{zhou2022preserve} & ARXIV'22 & 8 & --- & 82.3\\
\rowcolor{gray!10}ElrAr \cite{bai2023extreme} & IJCV'23 & 20 & --- & 56.2\\
\rowcolor{white!10}VARD \cite{lin2023self} & TIP'23 & 16 & --- & 72.6\\
\rowcolor{gray!10}STE-CapsNet \cite{feng2023spatial} & TOMM'23 & 4 & --- & 83.0\\
\rowcolor{white!10}STANet \cite{li2023spatio} & TCSVT'23 & 8 & --- & 87.7\\
\rowcolor{gray!10}SVFormer \cite{xing2023svformer} & CVPR'23 & 16 & --- & 86.7\\
\rowcolor{white!10}ActionHub \cite{zhou2024actionhub} & ARXIV'24 & 16 & --- & 51.2\\
\rowcolor{gray!10}Multi-Transforms \cite{vu2024self} & ICMEW'24 & 16 & --- & 63.2\\
\rowcolor{white!10}DTIF \cite{qian2024semi} & NURCMPT'24 & 8 & --- & 78.8\\
\rowcolor{blue!10}DVFL-Net (\textbf{Ours}) &  & 8 & 22M & \textbf{88.4}   \\
\hline
\end{tabular}
\label{tab:sota_ucf101}
\end{table}
\begin{table}[t]
\caption{\small Comparison of DVFL-Net with state-of-the-art human action recognition methods on the \textbf{HMDB51} dataset. Top-1 accuracy is used as the evaluation metric for comparison, where '---' indicates the unavailable information.}
\centering
\begin{tabular}{l|c|c|c|c}
\hline
\rowcolor{white!10}Method & Venue & Frames & Parameters &Top-1  \\
\hline
\rowcolor{gray!10}STS \cite{wang2021self} & TPAMI'21 & 16 & 9.6M & 34.4\\
\rowcolor{white!10}CVRL \cite{qian2021spatiotemporal} & CVPR'21 & 16 & 328M & 44.6\\
\rowcolor{gray!10}MFO \cite{qian2021enhancing} & ICCV'21 & 16 & --- & 47.6\\
\rowcolor{white!10}VideoMoCo \cite{pan2021videomoco} & CVPR'21 & 16 & 14.4M & 49.2\\
\rowcolor{gray!10}TCLR \cite{dave2022tclr} & CVIU'22 & 16 & --- & 55.4\\
\rowcolor{white!10}MEACI-Net \cite{li2022representation} & IJCAI'22 & 16 & 23.2M & 74.4\\
\rowcolor{gray!10}TCM \cite{liu2022motion} & TIP'22 & 16 & 49M & 77.5\\
\rowcolor{white!10}L2A \cite{gowda2022learn2augment} & ECCV'22 & 16 & --- & 47.1\\
\rowcolor{gray!10}LTG \cite{xiao2022learning} & CVPR'22 & 8 & --- & 49.7\\
\rowcolor{white!10}STM \cite{wang2022learning} & TPAMI'23 & 16 & 24M & 75.2\\
\rowcolor{gray!10}STANet \cite{li2023spatio} & TCSVT'23 & 8 & --- & 59.3\\
\rowcolor{white!10}ViT-ReT \cite{wensel2023vit} & ACCESS'23 & 20 & --- & \textit{78.4}\\
\rowcolor{gray!10}SVFormer \cite{xing2023svformer} & CVPR'23 & 16 & --- & 68.2\\
\rowcolor{white!10}StInNet \cite{jiang2024spatial} & TII'24 & 32 & 4.3M & 69.3\\
\rowcolor{gray!10}SRTN \cite{hussain2024hybrid} & TCE'24 & 30 & --- & 76.1\\
\rowcolor{white!10}CF-IIH \cite{liu2024knowledge} & TMM'24 & 16 & 186.M & 76.6\\
\rowcolor{blue!10}DVFL-Net (\textbf{Ours}) &  & 8 & 22M & \textbf{82.7} \\
\hline
\end{tabular}
\label{tab:sota_hmdb51}
\end{table}
\begin{table}[t]
\caption{\small Comparison of DVFL-Net with state-of-the-art human action recognition methods on the \textbf{SSV2} dataset. Top-1 accuracy is used as the evaluation metric for comparison, where '---' indicates the unavailable information.}
\centering
\begin{tabular}{l|c|c|c|c}
\hline
\rowcolor{white!10}Method & Venue & Frames & Parameters &Top-1  \\
\hline
\rowcolor{gray!10}TimeSformer-HR \cite{bertasius2021space} & ICML'21 & 96 & 121.4M & 62.5\\
\rowcolor{white!10}VidTR \cite{zhang2021vidtr} & ICCV'21 & 32 & - & 63.0\\
\rowcolor{gray!10}ViViT-L FE \cite{arnab2021vivit} & CVPR'21 & 32 & 86.7M & 65.9\\
\rowcolor{white!10}MFormer-L \cite{patrick2021keeping} & NeurIPS'21 & 8 & - & 68.1\\
\rowcolor{gray!10}MViTv1-B \cite{fan2021multiscale} & ICCV'21 & 32 & 36.6M & 67.6\\
\rowcolor{white!10}MTV-B \cite{yan2022multiview} & CVPR'22 & 16 & 224M & 67.6\\
\rowcolor{gray!10}Video-Swin-B \cite{liu2022video} & CVPR'22 & 32 & 88.8M & 69.6\\
\rowcolor{white!10}Uniformer-B \cite{li2022uniformer} & ICLR'22 & 32 & 49.8M & 70.4\\
\rowcolor{gray!10}MViTv2-B \cite{li2022mvitv2} & CVPR'22 & 16 & 51.1M & 70.5\\
\rowcolor{white!10}STW \cite{cai2023novel} & ACCESS'23 & 16 & - & 63.7 \\
\rowcolor{gray!10}D-TSM \cite{lee2023d} & ICUR'23 & 8 & - & 59.9 \\
\rowcolor{white!10}Video-FocalNet-B \cite{wasim2023video} & ICCV'23 & 16 & 157M & \textbf{71.1}\\
\rowcolor{gray!10}SAM \cite{wu2024scene} & CVIU'24 & 8 & - & 63.2\\
\rowcolor{white!10}STA+ \cite{li2024sta+} & DTPI'24 & - & 100 & 67.9\\
\rowcolor{gray!10}STASTA \cite{ting2024short} & ACCESS'24 & 16 & 49.3 & 68.5\\
\rowcolor{blue!10}DVFL-Net (\textbf{Ours}) &  & 8 & 22M & \underline{70.8} \\
\hline
\end{tabular}
\label{tab:sota_ssv2}
\end{table}
\noindent Similarly, on the Kinetics-400 dataset (Table \ref{tab:baseline_k400}), our VFL-Net achieves a Top-1 accuracy of 77.5\% and a Top-5 accuracy of 85.9\%, delivering competitive results despite its significantly smaller size compared to the baseline models. In comparison, Video-FocalNet Tiny, Small, and Base models reach Top-1 accuracies of 79.6\%, 81.3\%, and 83.1\%, with parameter counts of 49M, 88M, and 157M, respectively. Notably, VFL-Net achieves respectable accuracy with only 22M parameters, showcasing an effective balance between performance and computational efficiency, making it highly suitable for resource-constrained environments where model size is a crucial factor.

\subsection{VFL-Net Performance Evaluation Under KD Settings}
The results in Table~\ref{tab:kd_results} present the performance gains of our DVFL-Net on five action recognition datasets, including UCF50, UCF101, HMDB51, SSV2, and Kinetics-400 under knowledge distillation settings. For the UCF50 dataset, DVFL-Net shows a Top-1 accuracy of 86.6\%, marking a significant improvement of 5.2\% over the proposed baseline VFL-Net model. Similarly, the Top-5 accuracy reaches 95.2\%, a 3.8\% increase, demonstrating the effectiveness of our DVFL-Net architecture under the knowledge distillation settings in improving both the model’s overall accuracy and its ability to capture finer-grained actions.\\
\indent For the UCF101 dataset, our DVFL-Net achieves a Top-1 accuracy of 88.4\%, which is a notable 5.9\% gain, while its Top-5 accuracy rises to 96.1\%, an improvement of 2.4\%. The most substantial improvement is observed on the HMDB51 dataset, where DVFL-Net reaches a Top-1 accuracy of 82.7\%, reflecting a remarkable 11.1\% increase. The Top-5 accuracy also improves significantly, rising by 4.3\% to 92.8\%. For the SSV2 dataset, DVFL-Net achieves a Top-1 accuracy of 70.8\%, showing a significant improvement of 6.1\%, while its Top-5 accuracy rises to 78.7\%, reflecting a notable increase of 6.3\%. Similarly, for the Kinetics-400 dataset, DVFL-Net achieves a Top-1 accuracy of 83.1\%, with a 5.6\% gain, and a Top-5 accuracy of 89.3\%, showing an improvement of 3.9\%. These results demonstrate that knowledge distillation significantly boosts the model’s generalization capabilities on large-scale, diverse action classes, aligning with the observed improvements across UCF50, UCF101, HMDB51, and SSV2. The consistent accuracy gains across all datasets validate the robustness of the DVFL-Net architecture, especially in capturing complex spatio-temporal patterns.

\subsection{Comparison with State-of-The-Art}
When compared to state-of-the-art methods on the UCF50 dataset in Table~\ref{tab:sota_ucf50}, our DVFL-Net achieves a notable Top-1 accuracy of 86.6\%. This outperforms several recent methods, including AFC-BiGRU which achieved 85.8\%  while CLST-DNN and 3D-CNN obtained 80.8\% and 82.6\%, respectively. Despite these methods demonstrating competitive results, DVFL-Net delivers superior performance while maintaining a smaller model size, emphasizing its efficiency and effectiveness for human action recognition task. Additionally, other approaches such as ACE-BET and HDENN report lower Top-1 accuracy of 84.0\% and 75.0\%, highlighting the robustness of our DVFL-Net in both accuracy and efficiency.\\
\indent Similarly, on the UCF101 dataset in Table~\ref{tab:sota_ucf101}, our DVFL-Net continues to outperform several contemporary methods, achieving a Top-1 accuracy of 88.4\%. This result surpasses that of SVFormer (86.7\%) and STANet (87.7\%), as well as TCLR (85.4\%). These results indicate that DVFL-Net not only maintains high accuracy but also operates with lower parameter cost compared to the state-of-the-art HAR methods. Its performance across different benchmark datasets validates its robustness and ability to handle diverse action recognition scenarios.

\indent On the HMDB51 dataset in Table~\ref{tab:sota_hmdb51}, DVFL-Net demonstrates leading performance with a Top-1 accuracy of 82.7\%, notably surpassing several recent methods like STM (75.2\%) and STANet (59.3\%). While approaches such as MEACI-Net and TCM achieve Top-1 accuracies of 74.4\% and 77.5\%, respectively, DVFL-Net consistently maintains an accuracy advantage. These results highlight DVFL-Net's ability to deliver both high accuracy and model efficiency, making it a suitable choice for real-world scenarios where computational resources are constrained.

Similarly, our DVFL-Net achieves a Top-1 accuracy of 70.8\% on SSV2 dataset (Table \ref{tab:sota_ssv2}), surpassing many state-of-the-art methods, including MViTv2-B \cite{li2022mvitv2}, Uniformer-B \cite{li2022uniformer}, and Video-Swin-B \cite{liu2022video}, which achieve accuracies of 70.5\%, 70.4\%, and 69.6\%, respectively. Notably, our method achieves performance close to the best-performing Video-FocalNet-B \cite{wasim2023video}, which reaches 71.1\% accuracy, while using significantly fewer parameters (22M vs. 157M) and processing fewer frames (8 vs. 16). This underscores DVFL-Net’s efficiency, delivering near state-of-the-art accuracy with a more lightweight architecture and lower computational cost.

\begin{figure*}[t]
	\centering
	\includegraphics[width=1.0\linewidth]{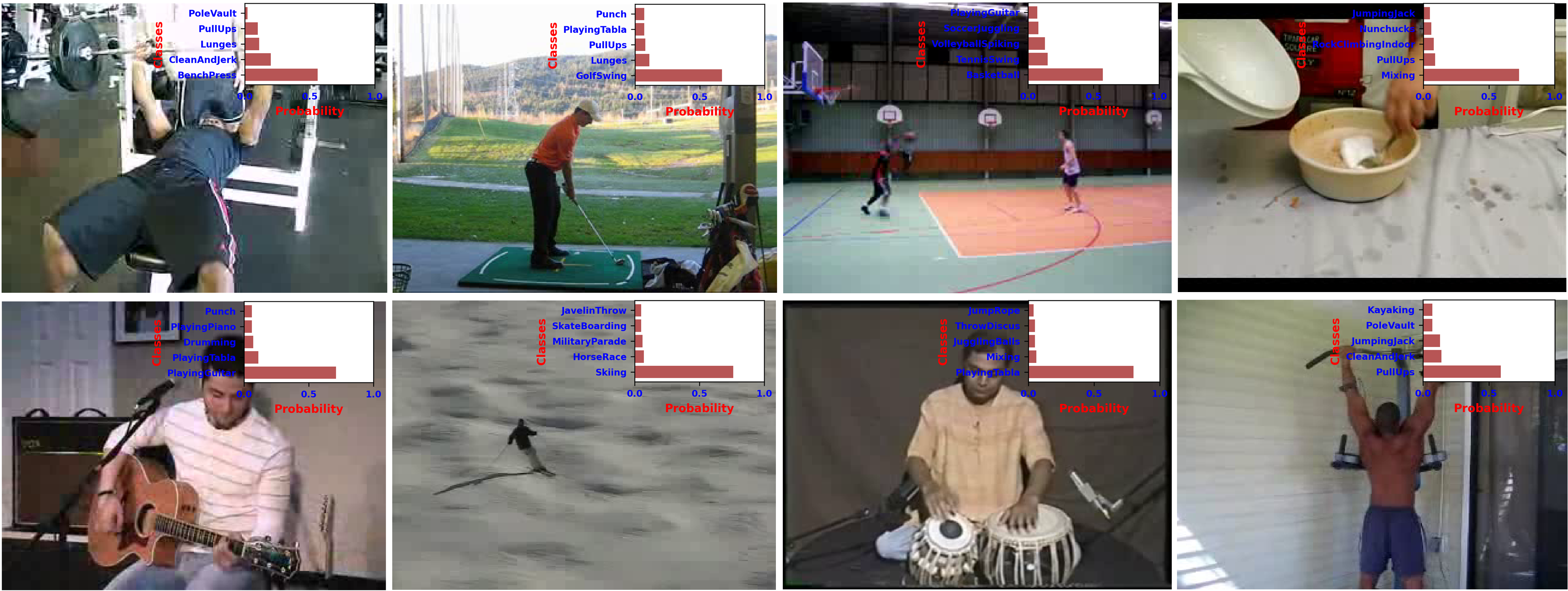}
	\caption{Visual illustration of VFLNet$_{kd}$'s \textbf{top-5 predictions} across eight distinct action videos sampled from various classes of UCF101 dataset, including \textbf{Bench Press}, \textbf{Golf Swing}, \textbf{Basketball}, \textbf{Mixing},\textbf{ Playing Guitar}, \textbf{Skiing}, \textbf{Playing Tabla}, and \textbf{Pullups}. Each video thumbnail is overlaid with a bar chart in the top-right corner, illustrating the model’s top-5 predicted classes and corresponding probabilities.}
	\label{fig:vis_pred}
\end{figure*}
\begin{table}[t]
\caption{\small Comparison of DVFL-Net with state-of-the-art human action recognition methods on the \textbf{Kinetics-400} dataset. Top-1 accuracy is used as the evaluation metric for comparison, where '---' indicates the unavailable information.}
\centering
\begin{tabular}{l|c|c|c|c}
\hline
\rowcolor{white!10}Method & Venue & Frames & Parameters &Top-1  \\
\hline
\rowcolor{gray!10}TEA \cite{li2020tea} & CVPR'20 & 8 & - & 76.1 \\
\rowcolor{white!10}VidTR-L \cite{zhang2021vidtr} & ICCV'21 & 32 & - & 79.1 \\
\rowcolor{gray!10}TimeSformer-L \cite{bertasius2021space} & ICML'21 & 96 & 121.4M & 80.7 \\
\rowcolor{white!10}MFormer-HR \cite{patrick2021keeping} & NeurIPS'21 & 8 & - & 81.1 \\
\rowcolor{gray!10}MViTv1-B \cite{fan2021multiscale} & ICCV'21 & 32 & 36.6 & 81.2 \\
\rowcolor{white!10}MoViNet-A6  \cite{kondratyuk2021movinets} & CVPR'21 & 120 & 31.4M & 81.5 \\
\rowcolor{gray!10}MTV-B \cite{yan2022multiview} & CVPR'22 & 16 & 224M & 81.8 \\
\rowcolor{white!10}Video-Swin-B \cite{liu2022video} & CVPR'22 & 32 & 88.8M & 82.7 \\
\rowcolor{gray!10}MViTv2-B \cite{li2022mvitv2} & CVPR'22 & 16 & 51.1M & 82.9 \\
\rowcolor{white!10}Uniformer-B \cite{li2022uniformer} & ICLR'22 & 32 & 49.8M & 83.0 \\
\rowcolor{gray!10}Video-FocalNet-B \cite{liu2022video} & ICCV'23 & 16 & 157M & \textbf{83.6} \\
\rowcolor{white!10}SVFormer \cite{xing2023svformer} & CVPR'23 & 16 & - & 69.4 \\
\rowcolor{gray!10}TimeBalance \cite{dave2023timebalance} & CVPR'23 & 8 & - & 61.2 \\
\rowcolor{white!10}DMTNet \cite{zhang2023dilated} & APPSCI’23 & 8 & - & 75.9 \\
\rowcolor{gray!10}SSAR \cite{wang2024learning} & ARXIV’24 & 8 & - & 69.9 \\
\rowcolor{white!10}DSFNet \cite{sun2024discriminative} & TMCCA'24 & 16 & - & 78.6 \\
\rowcolor{gray!10}TDAM \cite{zhang2024temporal} & NNICE'24 & 8 & 33M & 74.8 \\
\rowcolor{white!10}MDAF \cite{wang2024efficient} & ESWA'24 & 8 & - & 76.2 \\
\rowcolor{blue!10}DVFL-Net (\textbf{Ours}) &  & 8 & 22M & \underline{83.1} \\
\hline
\end{tabular}
\label{tab:sota_k400}
\end{table}
\indent In Table~\ref{tab:sota_k400}, our DVFL-Net achieves a Top-1 accuracy of 83.1\% on the Kinetics-400 dataset, outperforming most existing methods including MViTv2-B \cite{li2022mvitv2}, Uniformer-B \cite{li2022uniformer}, and Video-Swin-B \cite{liu2022video} having accuracies of 83.0\%, 82.9\%, and 82.7\%, respectively. It is worth noting that, despite having only 22M parameters, our DVFL-Net delivers performance comparable to the best-performing method, Video-FocalNet-B \cite{wasim2023video}, which achieves a Top-1 accuracy of 83.6\% but requires 157M parameters, nearly 7$\times$ more than our DVFL-Net model. Overall, these comparative analysis validate the DVFL-Net’s ability to deliver near state-of-the-art accuracy with a more lightweight architecture, making it a practical solution for real-world applications where computational efficiency is critical.
\begin{figure}[t]
	\centering
	\includegraphics[width=1.0\linewidth]{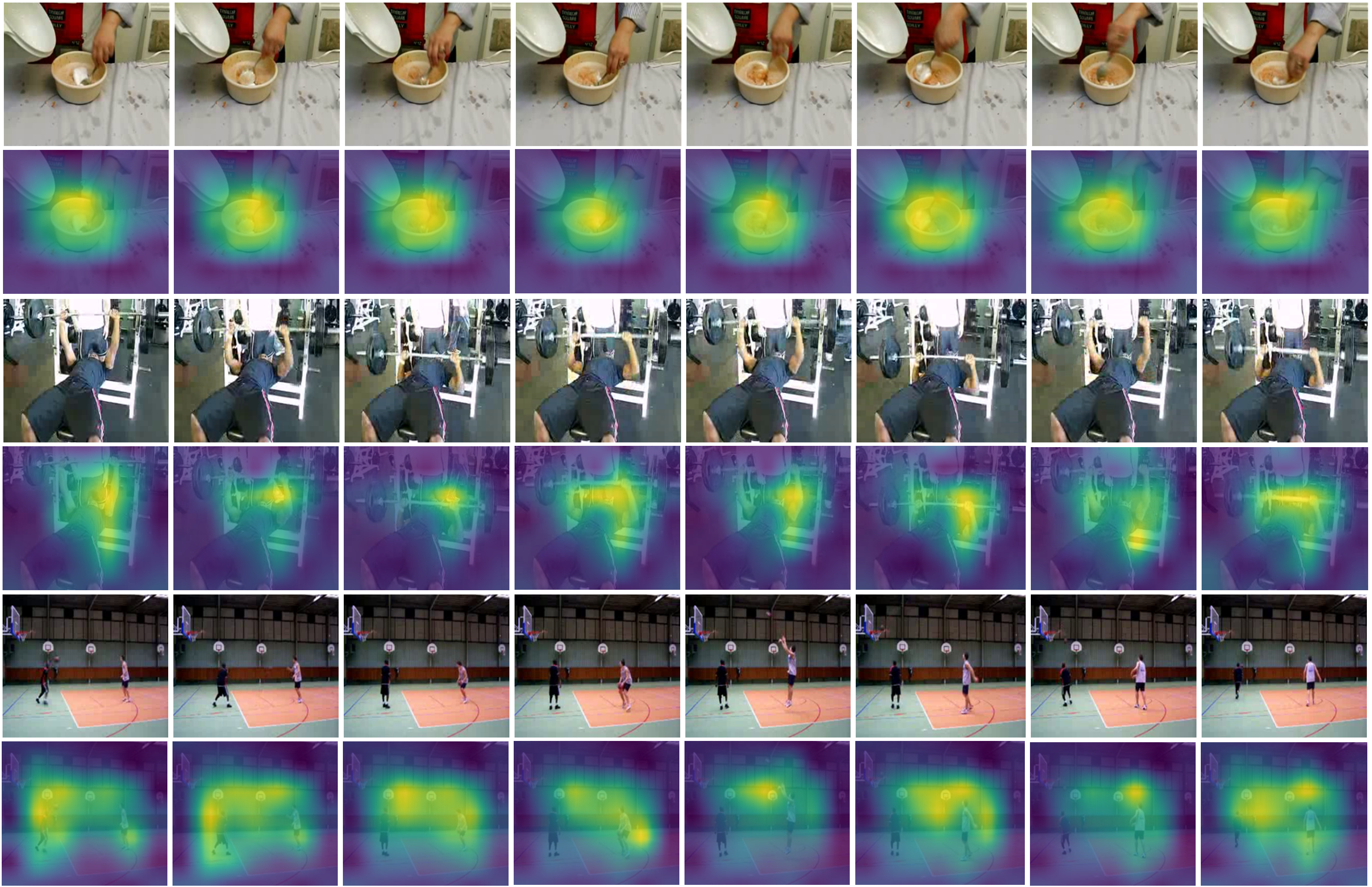}
	\caption{Visual illustration of \textbf{spatio-temporal focal modulation} of three sample videos (i.e., \textbf{Mixing}, \textbf{Bench Press}, and \textbf{Basketball}) from UCF101 dataset. The brighter regions in yellow and green correspond to areas of significant action, reflecting where our DVFL-Net focuses its attention. These visualizations illustrate how DVFL-Net captures and interprets spatio-temporal dynamics, effectively identifying the salient regions crucial for understanding the actions or objects present within each scene.}
	\label{fig:spatio-temp-focal}
\end{figure}
\subsection{Qualitative Results}   
To evaluate the qualitative performance of our DVFL-Net, we conducted experiments on eight video samples from distinct action categories in the UCF101 dataset, including Bench Press, Golf Swing, Basketball, Mixing, Playing Guitar, Skiing, Playing Tabla, and Pullups, as illustrated in Figure~\ref{fig:vis_pred}. For each video, the bar chart in the top-right corner of the thumbnail presents the top-5 predicted classes along with their corresponding probabilities. These results highlight the robustness of the proposed framework in discriminating between actions with similar visual or motion features, as evident from the high-confidence predictions in each sample. These visualizations validate the DVFL-Net's effectiveness in capturing both spatial and temporal cues essential for precise video-based action recognition.\\
\indent Further, we inspected the spatio-temporal focal modulation of our DVFL-Net on three videos sampled from different action categories of the UCF101 dataset. The obtained spatio-temporal focal modulation maps are visualized in Figure~\ref{fig:spatio-temp-focal}, highlighting the model’s ability to dynamically focus on critical regions across both spatial and temporal dimensions, effectively localizing action-relevant cues while suppressing irrelevant background information. The visualizations emphasize the model's capacity to capture and interpret complex motion patterns and object interactions, validating its robustness in spatio-temporal action recognition tasks.

\begin{figure}[t]
	\centering
	\includegraphics[width=\linewidth]{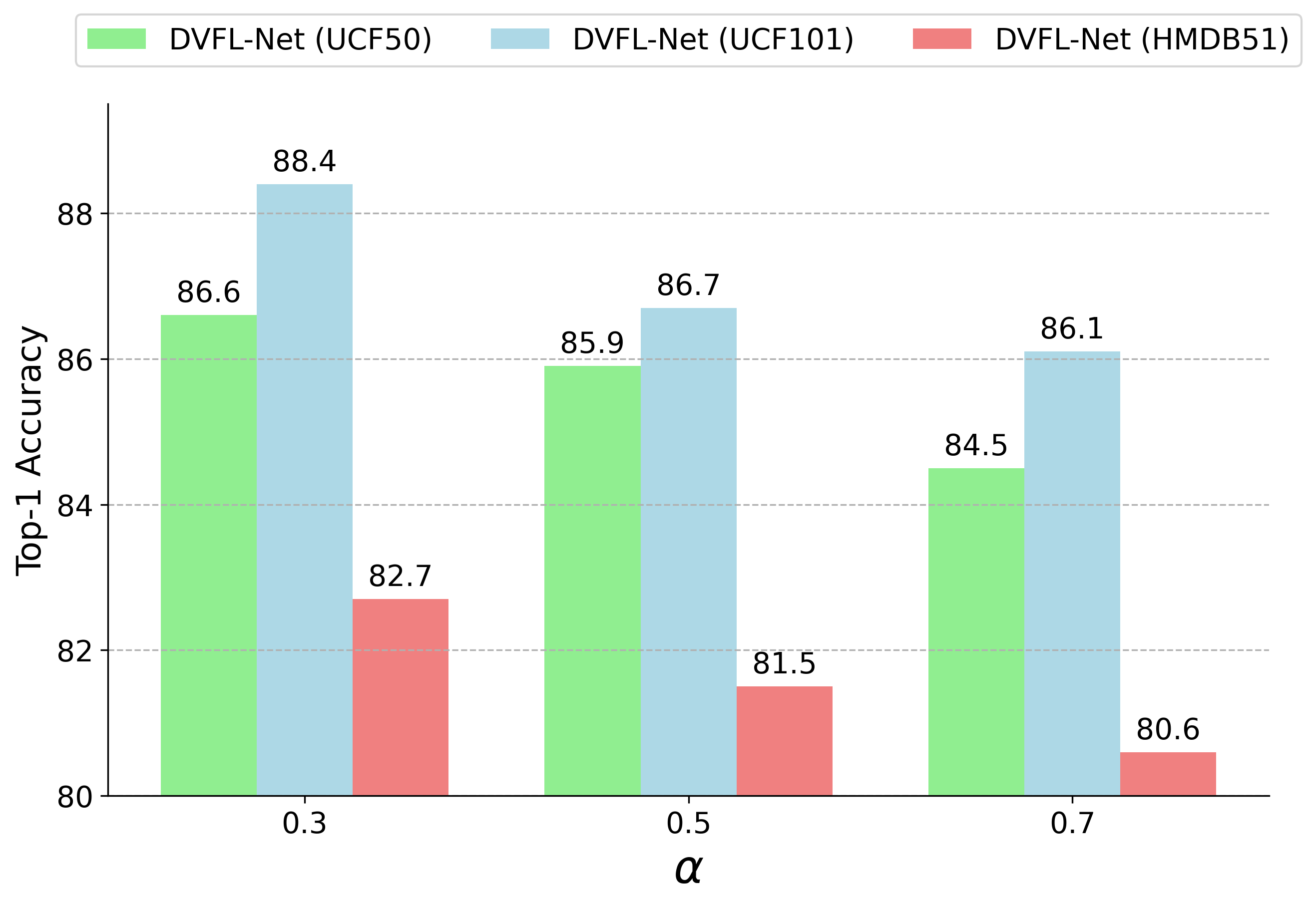}
	\caption{\small \textbf{Impact} of $\alpha$ on Top-1 accuracy of DVFL-Net across UCF101, UCF50, and HMDB51 datasets. The ablation study highlights performance variations with different $\alpha$ values.}
	\label{fig:alpha_ablation}
\end{figure}
\begin{figure}[t]
	\centering
	\includegraphics[width=\linewidth]{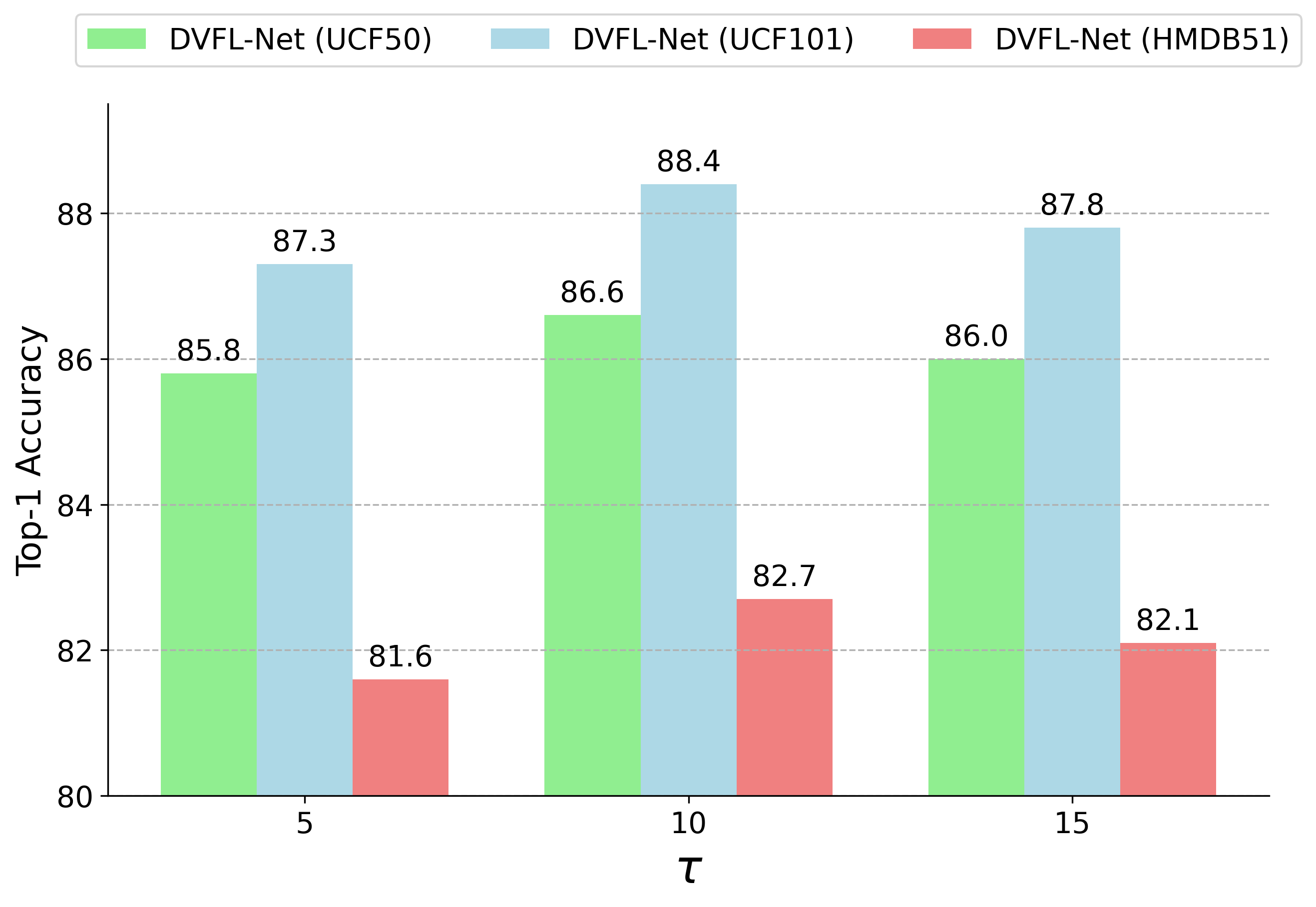}
	\caption{\small \textbf{Impact} of $\tau$ on Top-1 accuracy of VFL-Net across UCF101, UCF50, and HMDB51 datasets. The ablation study highlights performance variations with different $\tau$ values.}
	\label{fig:temp_ablation}
\end{figure}
\begin{figure}[t]
	\centering
	\includegraphics[width=\linewidth]{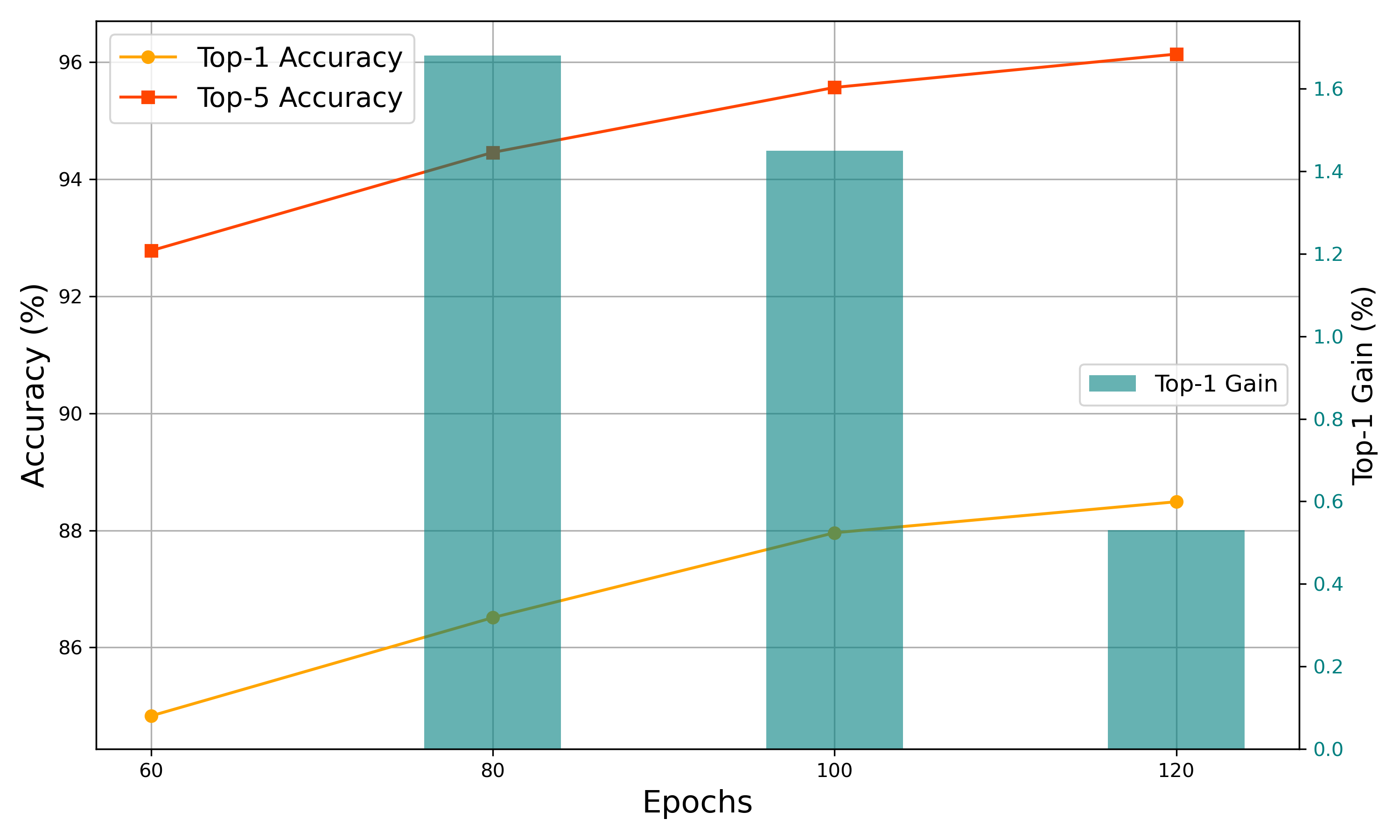}
	\caption{\small \textbf{Impact} of training duration on model's performance in knowledge distillation setting using UCF101 dataset.}
	\label{fig:kd_dur_ablation}
\end{figure}
\subsection{Ablation Study}      
To evaluate the impact of varying values of hyperparameters, specifically alpha and temperature, on the performance of knowledge distillation, we perform comprehensive ablation studies across the UCF50, UCF101, and HMDB51 datasets.
\subsubsection{Impact of Different Alpha Values}
Figure~\ref{fig:alpha_ablation} demonstrates the impact of different values of the hyperparameter alpha $\alpha$ on Top-1 accuracy across three datasets: UCF101, UCF50, and HMDB51. For the UCF101 dataset, the highest Top-1 accuracy is achieved when $\alpha$ = 0.3, reaching around 88.5\%. This indicates that at $\alpha$ = 0.3, there is an optimal balance between the knowledge transferred from the teacher model and the learning capacity of the student model, resulting in best Top-1 accuracy. As $\alpha$ increases beyond 0.3, a slight decrease in Top-1 accuracy is observed (such as $\alpha$ = 0.5 $\rightarrow$ 86.7\% and $\alpha$ = 0.7 $\rightarrow$ 86.1\%), suggesting that too much emphasis on the teacher's output can reduce the effectiveness of knowledge distillation.\\
\indent Similarly, on the UCF50 dataset, the best Top-1 accuracy of 86.6\% is achieved at $\alpha$ = 0.3, with a noticeable improvement in accuracy compared to other values of $\alpha$ (i.e., 0.5 and 0.7). For the HMDB51 dataset, performance trend is consistent with the highest Top-1 accuracy of 82.7\% at $\alpha$ = 0.3. Across each dataset used in this work, $\alpha$ = 0.3 yields the best results, highlighting that a moderate contribution from the teacher model leads to optimal learning during knowledge distillation.

\begin{figure*}[t]
	\centering
	\includegraphics[width=1.0\linewidth]{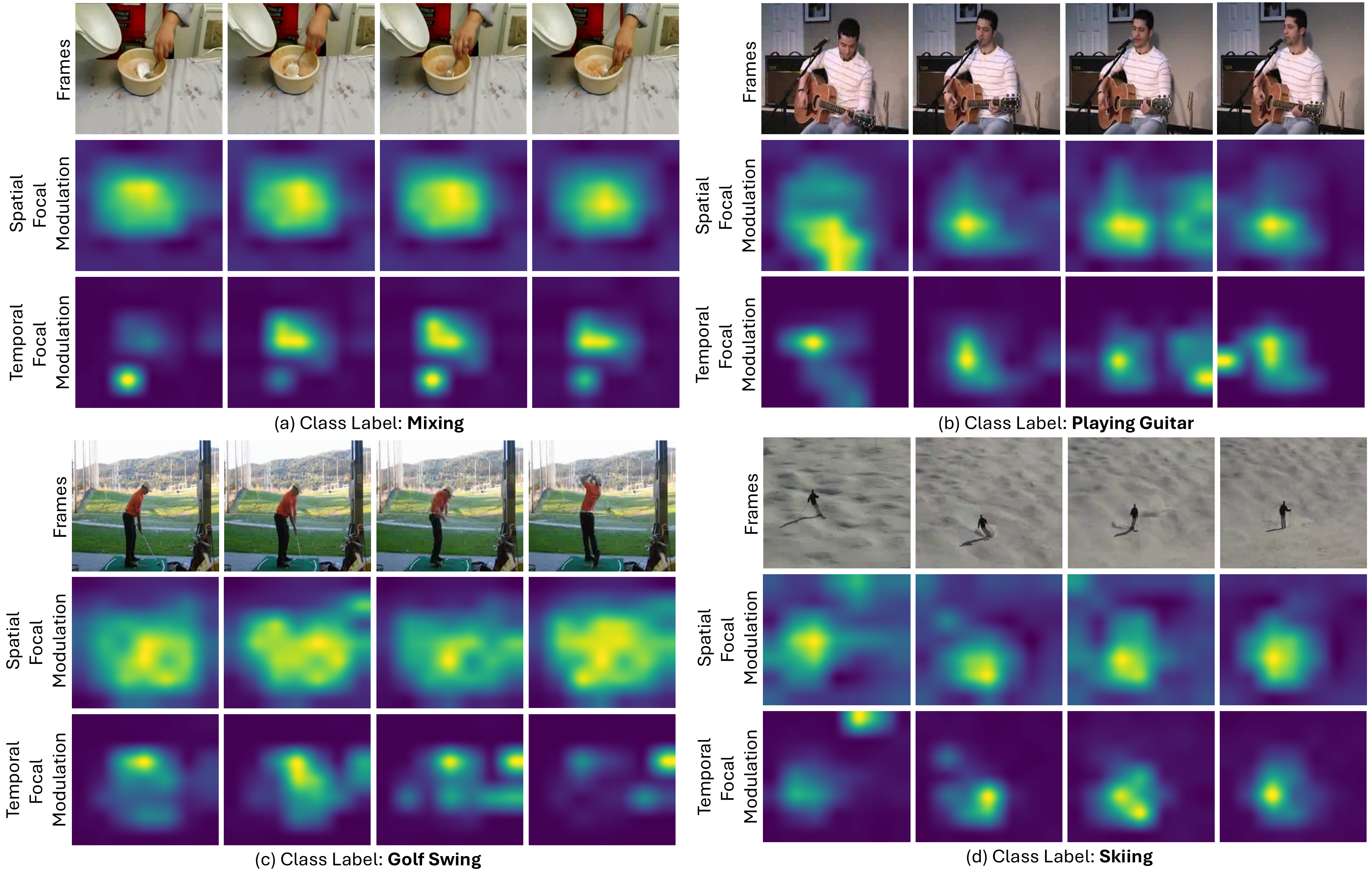}
	\caption{ Visual illustration of the spatial and temporal focal modulation achieved by our DVFL-Net on sample videos from the UCF101 dataset. The results demonstrate how the temporal modulator consistently captures global motion patterns across frames, effectively handling camera movement and small regions of interest. While, the spatial modulator dynamically adjusts focus across frames to capture local variations, emphasizing critical regions for accurate action understanding.}
	\label{fig:indi_modul}
\end{figure*}

\subsubsection{Impact of Different Temperature Values}
Figure~\ref{fig:temp_ablation} illustrates the impact of varying temperature $\tau$ values on Top-1 accuracy across three datasets, including UCF101, UCF50, and HMDB51, evaluated using the DVFL-Net model. The temperature parameter $\tau$ plays a crucial role in adjusting the softening of output probabilities of teacher and student model during knowledge distillation. Temperature $\tau$ smooth the probability distribution, enabling the student model to better capture the teacher model’s predictions. As shown in Figure~\ref{fig:temp_ablation}, Top-1 accuracy fluctuates with changes in $\tau$, indicating the sensitivity of model performance to this hyperparameter.\\
\indent The optimal performance across all three datasets is consistently observed at $\tau$ = 10. Specifically, UCF101 achieves a peak accuracy of ~88.4\%, UCF50 reaches ~86.6\%, and HMDB51 attains ~82.7\% under this ablation settings. As $\tau$ deviates from 10, the accuracy either remains stable or experiences a slight decline, indicating that $\tau$ = 10 is the optimal choice across datasets. Although there are minor differences in how each dataset responds to changes in $\tau$, this value offers the best trade-off for effective knowledge distillation in DVFL-Net.

\subsubsection{Visual Inspection of Spatio-Temporal Focal Modulation} To better understand the impact of individual focal modulation i,e., spatial focal modulation and temporal focal modulation, we thoroughly inspect them separately on four video samples from different action categories of the UCF101 dataset, as shown in Figure~\ref{fig:indi_modul}. As depicted in Figure~\ref{fig:indi_modul}, the spatio-temporal focal modulation of our DVFL-Net effectively captures salient regions and critical dynamics within the video, concentrating on features most relevant to the action recognition task. It is worth noticing that, the spatial focal modulator dynamically adapts to local spatial variations within individual frames, whereas the temporal focal modulator focuses on the global regions across frames, emphasizing areas with significant motion activity.

\subsubsection{Impact of Training Durability on KD Performance}
To assess the impact of training duration on model performance in a knowledge distillation setting, we trained our model on the UCF101 dataset using four different epoch counts: 60, 80, 100, and 120. The results, presented in Figure~\ref{fig:kd_dur_ablation}, demonstrate a consistent improvement in both Top-1 and Top-5 accuracy as the number of training epochs increases from 60 to 120. Specifically, the Top-1 accuracy improves from 85.0\% to 88.49\%, while the Top-5 accuracy increases from 92.78\% to 96.14\%, indicating that prolonged training allows the student model to better capture the soft-target distribution provided by the teacher. However, the bar plot, which represents the incremental Top-1 accuracy gain across consecutive training stages, reveals diminishing returns over time. The most significant improvement 1.6\% is observed between 60 and 80 epochs, followed by smaller gains of approximately 1.4\% and 0.6\% in subsequent intervals. This trend suggests that, although prolonged training continues to improve performance, the benefits become marginal beyond a certain point, specifically after 120 epochs in our case. Thus, while training for more epochs can lead to better KD performance, the diminishing accuracy gains raise important considerations about the trade-off between model improvement and computational cost. Therefore, identifying the optimal training duration is crucial for achieving efficiency without compromising performance.

\subsection{Model Complexity and Memory Utilization}
\begin{figure}[t]
	\centering
	\includegraphics[width=\linewidth]{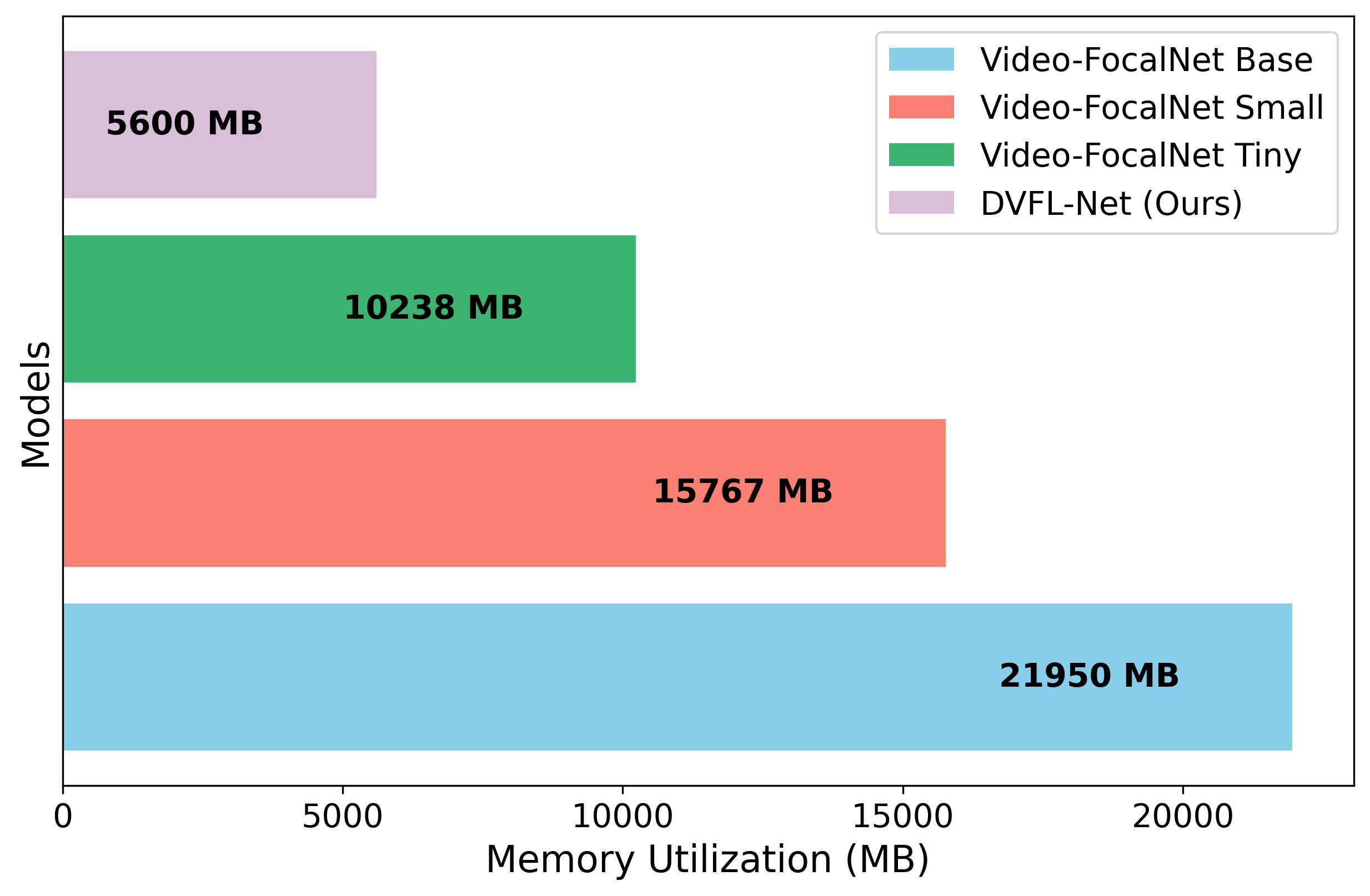}
	\caption{\small \textbf{Memory Utilization} across models on UCF101 dataset. Our DVFL-Net achieves the lowest memory usage and computational cost, demonstrating efficiency improvements over Video-FocalNet variants.}
	\label{fig:mem_util}
\end{figure}

We further extend our comparison of our DVFL-Net with other baseline models within the Video-FocalNet family, focusing on key metrics such as memory utilization and GFLOPs, as illustrated in Figure~\ref{fig:mem_util}. Our analysis reveals that DVFL-Net achieves the most efficient memory usage, consuming only 5600 MB per epoch, which is significantly lower compared to other models in the Video-FocalNet family. In contrast, Video-FocalNet Base shows the highest memory consumption, reaching up to 21950 MB per epoch. This substantial difference highlights the efficiency of our DVFL-Net model in terms of memory management, which is crucial for scaling models to larger datasets or deploying them on devices with limited resources.\\
\indent Following Video-FocalNet Base, the VFL-Net Small model is the second highest in memory consumption, utilizing 15767 MB per epoch. Despite its relatively smaller architecture, the increase in memory usage indicates a trade-off between model size and computational efficiency, which is an important consideration for model selection based on deployment scenarios. Interestingly, the VFL-Net Tiny model, although having the least number of parameters in the entire Video-FocalNet family, still consumes 2$\times$ more memory (10238 MB) than our DVFL-Net model. These observations suggests that our knowledge distillation strategy not only reduces the parameter count but also optimizes memory usage effectively.\\
\indent Besides, we also compare the model complexity of our DVFL-Net with other baseline models of the Video-FocalNet family across five datasets: UCF50, UCF101, HMDB51, SSV2, and Kinetics-400 as listed in Table~\ref{tab:model_complexity}. The complexities of these models are compared in terms of total training time (in hours), time per epoch (in minutes), GFLOPs (Giga Floating Point Operations per second), and the number of parameters (in millions). For the UCF50 dataset, DVFL-Net demonstrates significant efficiency, taking only 2.36 hours with 1.18 minutes per epoch, while utilizing just 27 GFLOPs and 22M parameters. In contrast, the Video-FocalNet Base model requires a total of 4.86 hours of training time with 2.43 minutes per epoch, utilizing 220 GFLOPs and 157M parameters. This shows that DVFL-Net achieves reduced computational complexity while maintaining its performance capabilities.
\begin{table*}[t]
\caption{\small \textbf{Complexity comparison} of DVFL-Net and baseline Video-FocalNet models across \textbf{UCF50}, \textbf{UCF101}, \textbf{HMDB51}, \textbf{SSV2}, and \textbf{Kinetics-400} datasets. The table reports total training time (\textbf{hours}), time per epoch (\textbf{minutes}), GFLOPs, and the number of parameters (\textbf{millions}) for each model configuration.}
\centering
\begin{tabular}{l|c|c|c|c|c}
\hline
\rowcolor{white!10}Model & Dataset & Total Training Time (Hours) & Time/Epoch (Minutes) & GFLOPs & Parameters\\
\hline
\cellcolor{gray!10}Video-FocalNet Tiny \cite{wasim2023video} & \multirow{4}{*}{\rotatebox[origin=c]{0}{UCF50}} & \cellcolor{gray!10}2.55 & \cellcolor{gray!10}1.28 & \cellcolor{gray!10}63 & \cellcolor{gray!10}49M \\ 
\cellcolor{white!10}Video-FocalNet Small \cite{wasim2023video} & & \cellcolor{white!10}2.86 & \cellcolor{white!10}1.44 & \cellcolor{white!10}124 & \cellcolor{white!10}88M   \\
\cellcolor{gray!10}Video-FocalNet Base \cite{wasim2023video} &  & \cellcolor{gray!10}4.86 & \cellcolor{gray!10}2.43 & \cellcolor{gray!10}220  & \cellcolor{gray!10}157M \\
\cellcolor{blue!10}DVFL-Net (\textbf{Ours}) &  &  \cellcolor{blue!10}2.36 & \cellcolor{blue!10}1.18 & \cellcolor{blue!10}27 & \cellcolor{blue!10}22M \\ 
\hline
\cellcolor{gray!10}Video-FocalNet Tiny \cite{wasim2023video} & \multirow{4}{*}{\rotatebox[origin=c]{0}{UCF101}} & \cellcolor{gray!10}5.26 & \cellcolor{gray!10}2.63 & \cellcolor{gray!10}63 & \cellcolor{gray!10}49M\\
\cellcolor{white!10}Video-FocalNet Small \cite{wasim2023video} &  & \cellcolor{white!10}5.75 & \cellcolor{white!10}2.87 & \cellcolor{white!10}124 & \cellcolor{white!10}88M  \\
\cellcolor{gray!10}Video-FocalNet Base \cite{wasim2023video} &  & \cellcolor{gray!10}5.78 & \cellcolor{gray!10}2.89 & \cellcolor{gray!10}220 & \cellcolor{gray!10}157M \\
\cellcolor{blue!10}DVFL-Net (\textbf{Ours}) &  & \cellcolor{blue!10}4.98 & \cellcolor{blue!10}2.49 & \cellcolor{blue!10}27 & \cellcolor{blue!10}22M \\ 
\hline
\cellcolor{gray!10}Video-FocalNet Tiny \cite{wasim2023video} & \multirow{4}{*}{\rotatebox[origin=c]{0}{HMDB51}} & \cellcolor{gray!10}2.95 & \cellcolor{gray!10}1.47 & \cellcolor{gray!10}63 & \cellcolor{gray!10}49M \\
\cellcolor{white!10}Video-FocalNet Small \cite{wasim2023video} &  & \cellcolor{white!10}4.93 & \cellcolor{white!10}2.47 & \cellcolor{white!10}124 & \cellcolor{white!10}88M  \\
\cellcolor{gray!10}Video-FocalNet Base \cite{wasim2023video} &  & \cellcolor{gray!10}5.55 & \cellcolor{gray!10}2.77 & \cellcolor{gray!10}220 & \cellcolor{gray!10}157M  \\
\cellcolor{blue!10}DVFL-Net (\textbf{Ours}) &  & \cellcolor{blue!10}2.40 & \cellcolor{blue!10}1.20 & \cellcolor{blue!10}27 & \cellcolor{blue!10}22M \\ 
\hline
\cellcolor{gray!10}Video-FocalNet Tiny \cite{wasim2023video} & \multirow{4}{*}{\rotatebox[origin=c]{0}{SSV2}} &\cellcolor{gray!10} 88.89 &\cellcolor{gray!10} 44.45 &\cellcolor{gray!10} 63 &\cellcolor{gray!10} 49M \\
\cellcolor{white!10}Video-FocalNet Small \cite{wasim2023video} &  &\cellcolor{white!10} 97.17 &\cellcolor{white!10} 48.50 &\cellcolor{white!10} 124 &\cellcolor{white!10} 88M  \\
\cellcolor{gray!10}Video-FocalNet Base \cite{wasim2023video} &  &\cellcolor{gray!10} 97.68 &\cellcolor{gray!10} 48.84 &\cellcolor{gray!10} 220 &\cellcolor{gray!10} 157M  \\
\cellcolor{blue!10}DVFL-Net (\textbf{Ours}) &  &\cellcolor{blue!10} 84.16 &\cellcolor{blue!10} 42.08 &\cellcolor{blue!10} 27 &\cellcolor{blue!10} 22M \\ 
\hline
\cellcolor{gray!10}Video-FocalNet Tiny \cite{wasim2023video} & \multirow{4}{*}{\rotatebox[origin=c]{0}{blue}{Kinetics-400}} &\cellcolor{gray!10} 123.92 &\cellcolor{gray!10} 61.96 &\cellcolor{gray!10} 63 &\cellcolor{gray!10} 49M \\ 
\cellcolor{white!10} Video-FocalNet Small \cite{wasim2023video} &  &\cellcolor{white!10} 135.47 &\cellcolor{white!10} 67.61 &\cellcolor{white!10} 124 &\cellcolor{white!10} 88M  \\
\cellcolor{gray!10}Video-FocalNet Base \cite{wasim2023video} &  &\cellcolor{gray!10} 136.17 &\cellcolor{gray!10} 68.08 &\cellcolor{gray!10} 220 &\cellcolor{gray!10} 157M  \\
\cellcolor{blue!10}DVFL-Net (\textbf{Ours}) &  &\cellcolor{blue!10} 117.32 &\cellcolor{blue!10} 58.66 &\cellcolor{blue!10} 27 &\cellcolor{blue!10} 22M \\ 
\hline
\end{tabular}
\label{tab:model_complexity}
\caption*{\footnotesize Note: All models are trained with a batch size of 8, and the number of frames is set to 8.}
\end{table*}
\vspace{3pt}
\indent Similarly, for the UCF101 and HMDB51 datasets, DVFL-Net consistently exhibits the lowest complexity among all models. For UCF101, DVFL-Net completes training in 4.98 hours and requires only 2.49 minutes per epoch, significantly lower than the Video-FocalNet Base model, which takes 5.78 hours and 2.89 minutes per epoch. Additionally, for the HMDB51 dataset, DVFL-Net’s training time is 2.40 hours with 1.20 minutes per epoch, highlighting its efficiency compared to other models. On the other hand, for large datasets such as SSV2 and Kinetics-400, our method achieves the lowest computational complexity in terms of training time. For instance, DVFL-Net completes training on the SSV2 dataset in just 84.16 hours, with each epoch taking 42.08 minutes. In comparison, Video-FocalNet Base requires 97.68 hours of training, with 48.84 minutes per epoch. Similarly, on the Kinetics-400 dataset, our DVFL-Net completes training in 117.32 hours, with each epoch taking 58.66 minutes, whereas Video-FocalNet Base takes 136.17 hours, with 68.08 minutes per epoch. Thus, the consistent reduction in training time, GFLOPs, and parameters across all datasets demonstrates that DVFL-Net is optimized for lower computational resources, making it an efficient alternative without sacrificing performance.

\section{Conclusion} \label{sec:conclusion}
In this work, we extensively explore various design configurations of the Video-FocalNet family architectures and propose a computationally efficient, lightweight DVFL-Net model for spatio-temporal action recognition. Our DVFL-Net model leverages a nano-scale spatio-temporal focal modulation mechanism, effectively capturing both short- and long-term dependencies to learn robust spatio-temporal representations. By using this mechanism, the model achieves hierarchical contextualization by combining spatial and temporal convolution and multiplication operations in parallel, optimizing computational efficiency. Furthermore, we employ forward KL divergence to transfer rich spatio-temporal knowledge from the larger Video-FocalNet Base (teacher) model, with 175 million parameters, to our lightweight VFL-Net model (student), which has only 22 million parameters. We thoroughly evaluate the performance of VFL-Net, both with and without forward KL divergence, against other baseline models from the Video-FocalNet family, including Video-FocalNet Base, Small, and Tiny. Notably, the distilled version of our VFL-Net, termed DVFL-Net, shows significant accuracy improvements of 5.2\%, 5.9\%, 11.1\%, 6.1\%, and 5.6\% on the UCF50, UCF101, HMDB51, SSV2, and Kinetics-400 datasets, respectively.\\
\indent Moreover, we compare the performance of DVFL-Net with recent state-of-the-art HAR methods. The comparative analysis highlights the superiority of DVFL-Net over existing methods across all evaluated datasets, demonstrating its optimal balance between performance and computational complexity. These results establish DVFL-Net as a highly efficient solution for human action recognition tasks.

\section*{Acknowledgment} 
This research was supported in part by the Air Force Office of Scientific Research (AFOSR) Contract Number FA9550-22-1-0040. The authors would like to acknowledge Dr. Erik Blasch from the Air Force Research Laboratory (AFRL) for his guidance and support on the project. Any opinions, findings, and conclusions or recommendations expressed in this material are those of the author(s) and do not necessarily reflect the views of the Air Force, the Air Force Research Laboratory (AFRL), and/or AFOSR.

\bibliographystyle{IEEEtran}
\bibliography{References}
\vskip -2\baselineskip plus -1fil
\begin{IEEEbiography}[{\includegraphics[width=2.0in,
height=1.3in,clip, keepaspectratio]{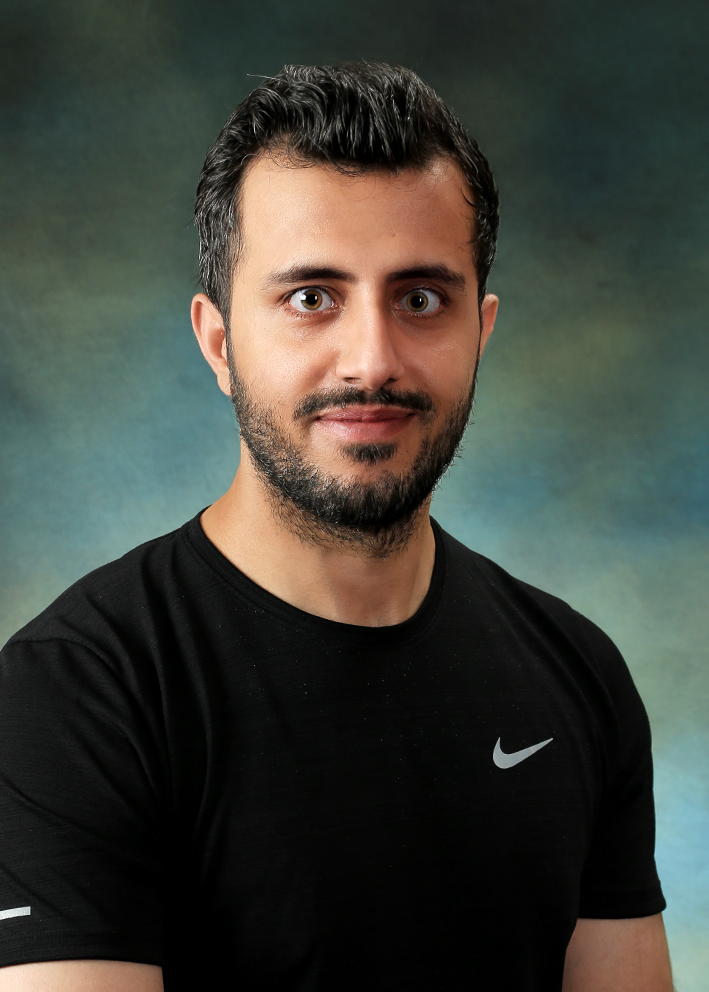}}]{Hayat Ullah} received his Bachelor's degree in Computer Science from Islamia College University, Peshawar, Pakistan, in 2018, and his Master's degree in Computer Science from Sejong University, Seoul, Republic of Korea, in 2021. He is currently pursing his Ph.D. in Computer Science at Florida Atlantic University, Boca Raton, FL, USA. He is also a Research Assistant with the Intelligent Systems, Computer Architecture, Analytics, and Security (ISCAAS) Laboratory, Florida Atlantic University, exclusively working on multi-model human actions modeling and activity recognition. He has published several articles in well-reputed journals, that include IEEE Internet of Things Journal and IEEE Transactions on Image Processing. His research interests include human action recognition, temporal action localization, knowledge distillation, adversarial robustness, and vision-language models for videos.
\end{IEEEbiography}

\vskip -2\baselineskip plus -1fil
\begin{IEEEbiography}[{ \includegraphics[width=2.0in,
height=1.3in,clip, keepaspectratio]{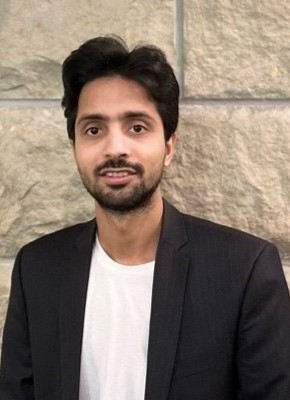}}]{Muhammad Ali Shafique} received the Bachelor of Science and master’s degrees from the University of Engineering and Technology, Lahore, in 2015 and 2017, respectively. He is currently pursuing the Ph.D. degree in Electrical and Computer Engineering at Kansas State University. He also holds the position of a Research Assistant with the ISCAAS Laboratory. His research interest includes design of efficient LLMs in resource-constrained systems. He is committed to conducting research in this area with the aim of advancing the current knowledge and understanding of these fields.
\end{IEEEbiography}

\vskip -2\baselineskip plus -1fil
\begin{IEEEbiography}[{ \includegraphics[width=2.0in,
height=1.3in,clip, keepaspectratio]{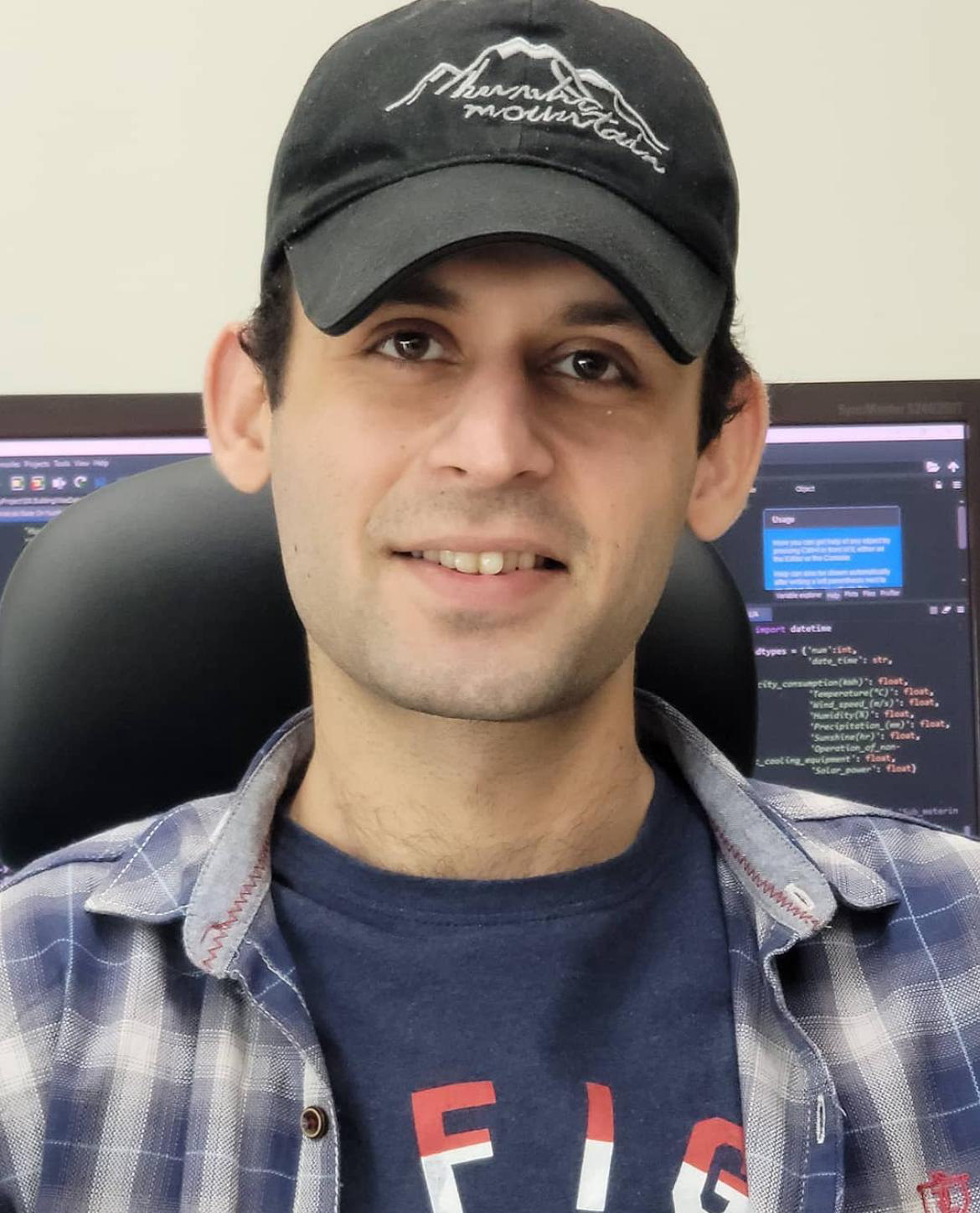}}]{Abbas Khan} received his Bachelor's degree in Computer Science from Islamia College Peshawar, Pakistan, and an M.S. degree in Software Convergence from Sejong University, Seoul, Republic of Korea. During his time at Sejong University, he worked as a Research Assistant at the Intelligent Media Laboratory (IMLab). Later, he joined the Department of Computer Vision as a Research Assistant at Muhammad Bin Zayed University of Artificial Intelligence, Abu Dhabi, UAE. Currently, he is pursuing a Ph.D. at Florida Atlantic University, Boca Raton, FL, USA, while also serving as a Research Assistant at the Intelligent Systems, Computer Architecture, Analytics, and Security (ISCAAS) Laboratory. His research interests span artificial intelligence and computer vision, focusing on Camouflaged Object Detection, Object Segmentation, Pattern Recognition, Active Learning, Vision-Language Models, Image Super-Resolution, and Image Defogging.
\end{IEEEbiography}

\vskip -2\baselineskip plus -1fil
\begin{IEEEbiography}[{\includegraphics[width=2.0in,
height=1.3in,clip, keepaspectratio]{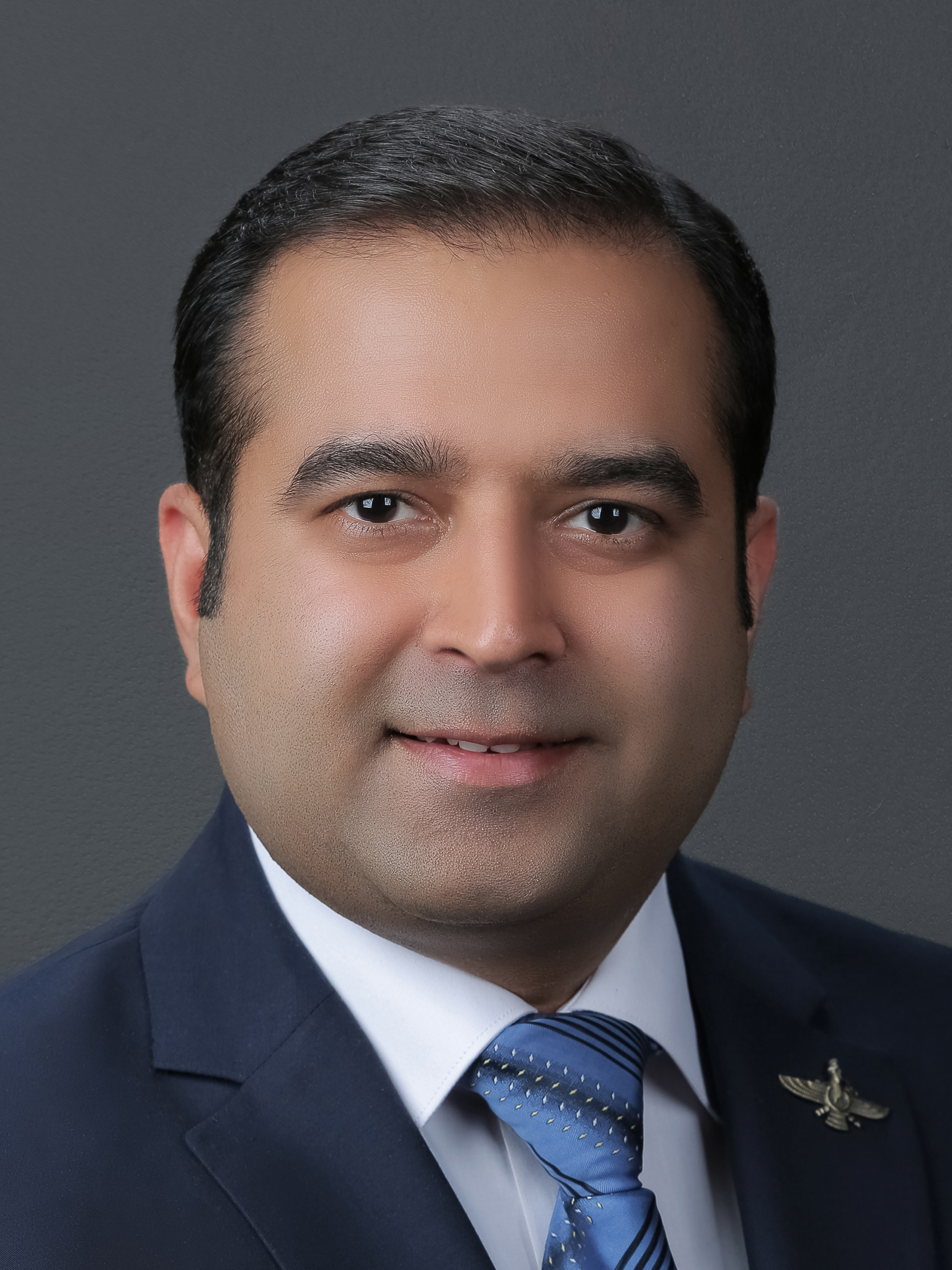}}]{Arslan Munir} (M'09,
SM'17) received his M.A.Sc. degree in electrical and computer engineering (ECE) from the University of British Columbia, Vancouver, Canada, in 2007, and his Ph.D. degree in ECE from the University of Florida, Gainesville, FL, USA, in 2012. From 2007 to 2008, he worked as a Software Development Engineer at the Embedded Systems Division, Mentor Graphics Corporation. He was a Postdoctoral Research Associate with the ECE Department at Rice University, Houston, TX, USA, from May 2012 to June 2014. He is currently an Associate Professor in the Department of Electrical Engineering and Computer Science at Florida Atlantic University. He is also an Adjunct Associate Professor of Computer Science and the University Outstanding Scholar at Kansas State University. 

Munir's current research interests include embedded and cyber-physical systems, secure and trustworthy systems, parallel computing, artificial intelligence, and computer vision. He has received many academic awards, including the Doctoral Fellowship from the Natural Sciences and Engineering Research Council (NSERC) of Canada. He earned gold medals for best performance in electrical engineering and gold medals and academic roll of honor for securing rank one in pre-engineering provincial examinations (out of approximately 300,000
candidates).
\end{IEEEbiography}


\end{document}